\documentclass[anonymous]{isprs} 
\usepackage{setspace}
\usepackage{geometry} 
\usepackage{epstopdf}
\usepackage[labelsep=period]{caption}  
\usepackage[british]{babel} 
\usepackage[hang]{footmisc}
\usepackage{natbib}
\usepackage{subcaption}
\usepackage{siunitx}
\usepackage{graphicx}
\usepackage{arydshln} 
\usepackage{booktabs}
\usepackage{xcolor}
\usepackage{multirow}
\usepackage{adjustbox}
\usepackage{enumitem}
\usepackage{svg}
\usepackage[hang]{footmisc}
\usepackage{url}
\usepackage{import}
\usepackage{graphicx}
\usepackage{bm}
\usepackage{svg}
\usepackage[table]{xcolor}
\usepackage{xcolor}
\usepackage{comment}
\usepackage{amsmath}
\usepackage{amssymb}
\usepackage{float}

\newcommand{\Point}[1]{{\ensuremath{\textbf{\textit{#1}}}}}

\newcommand{\Matrix}[1]{{\ensuremath{\textnormal{\textbf{\textit{#1}}}}}}
\newcommand{\TMatrix}[3]{{{}^{\mathrm{#1}}\Matrix{#2}_{\mathrm{#3}}}}
\usepackage{pifont}

\newcommand{\Unit}[1]{\thinspace #1}
\usepackage{todonotes}

\begin{document}

\title{MVM-IOD: An Industrial Object-Centric Benchmark Dataset \\ for the Evaluation of 3D Reconstruction Methods}
\date{}


%
\author{ \begin{tabular}{c}
Robert Langend\"orfer, Markus Hillemann, Markus Ulrich\\
\end{tabular}
 }


\address{
    \begin{tabular}{c}
    Machine Vision Metrology, Institute of Photogrammetry and Remote Sensing, Karlsruhe Institute of Technology, Germany - \\(robert.langendoerfer, markus.hillemann, markus.ulrich)@kit.edu\\
    \end{tabular}
  }



\abstract{3D object reconstruction, and camera pose estimation in industrial applications are challenging tasks, as errors are costly while the computation time is often limited. The complexity of typical industrial objects further complicates these tasks. Most of the existing datasets in this context do not depict realistic industrial scenarios. Therefore, we introduce the Machine Vision Metrology Industrial Object Dataset (MVM-IOD). Images of typical industrial objects are captured systematically, by moving a camera, mounted at the end effector of an industrial robot arm, on a hemisphere around the objects. MVM-IOD contains reference camera poses and reference 3D point clouds, the acquired RGB images of 9 objects and 2 background choices resulting in 18 scenes, which allows evaluation of all image based methods that compute a 3D reconstruction, camera poses, or novel views of a scene. Based on MVM-IOD, we extensively evaluate current SOTA 3D reconstruction and camera pose estimation methods, such as Structure from Motion, Multi-View Stereo, recent feed forward methods (Visual Geometry Grounded Transformer, $\pi^3$), and 2D Gaussian Splatting and report our findings as a baseline for future research. The experiments show that capture setups like ours generate out-of distribution images for feed forward methods, leading to suboptimal point clouds and camera poses. However, these out-of distribution images can be shifted closer to the training distribution by applying simple preprocessing steps. Consequently, in certain industrial applications, feed forward methods should be used with caution. The data and code are accessible via our project page: \url{https://pimpilimpo.github.io/projects/Industrial_objects/}

}

\maketitle

\sloppy
\section{Introduction}
\label{sec:Introduction}
In machine vision applications, such as anomaly detection, object pose estimation, and object grasping, an accurate 3D reconstruction and accurate camera poses are crucial \citep{steger2018machine}. The lack of texture on the surface of most industrial objects makes these tasks harder with data captured with cameras \citep{karami2021investigating}. Furthermore, time is often a limiting factor, as many applications need to be performed in real time for cost efficiency \citep{1242127}. In principle, there are various methods available to solve these tasks. However, to determine which methods work best on objects with certain properties (e.g., geometry, texture, material) a benchmark dataset is needed. Only then, an objective evaluation of current 3D reconstruction and camera pose estimation methods can be achieved. While there are existing datasets containing images and a corresponding 3D point cloud reference \citep{aanaes2016large,yan2023nerfbk}, they are lacking critical aspects that make them unsuitable for the evaluation of image-based 3D reconstruction and camera pose estimation in industrial settings. 

In this paper, we introduce MVM-IOD, a novel dataset that can be used in context of several typical computer and machine vision tasks, such as Multi-View Stereo (MVS), Novel View Synthesis (NVS), hand--eye calibration, and camera pose estimation. The dataset contains nine different objects (see Figure \ref{fig:objects_img}), which are typical for industrial applications and at the same time challenging for computer and machine vision methods. These challenges include lack of texture, transparent or non-Lambertian surfaces, very fine geometric structures, and different kinds of symmetries.
The dataset provides $300$ RGB images for each object, captured by an industrial camera mounted to the end effector of a high-precision industrial robot arm. Further, it provides accurate robot and camera poses for each of the captured images. For the camera poses, the dataset contains two sets: The first one is calculated based on the queried robot poses and the hand--eye pose, which was obtained by hand--eye calibration. The data and the results of the hand--eye calibration are also provided. The second one, which serves as reference, is calculated based on a calibration plate and spatial resection. This also requires the interior orientation of the camera, which was determined by camera calibration and is provided as well. Moreover, the dataset contains an accurate reference for the 3D point cloud of the objects, which has been obtained with a highly-accurate industrial line scanner.

We choose COLMAP as long-term standard method for Structure-from-Motion (SfM) \citep{schoenberger2016sfm} and MVS \citep{schonberger2016pixelwise} to generate a baseline for evaluating MVM-IOD on the task of pose estimation and 3D reconstruction. The time-critical aspect of most industrial applications motivates us to evaluate recent feed-forward visual geometry reconstruction methods such as Visual Grounded Geometry Transformer (VGGT) \citep{wang2025vggt} and subsequent work such as $\pi^3$ \citep{wang2025pi} as well, as they can solve the camera pose estimation and 3D reconstruction task at inference time within a few seconds. Furthermore, the robustness of these methods is evaluated, which is especially important in industrial applications where errors are costly. To show that the task of NVS can also be evaluated on this dataset, we incorporate 2D Gaussian Splatting (2DGS) \citep{huang20242d} making a comparison to Gaussian Splatting methods and other NVS algorithms possible, though we focus on its generated 3D reconstruction. We compare the estimated camera poses of VGGT, $\pi^3$ and SfM to the reference poses in order to make conclusions about the robustness of these methods in industrial settings. Furthermore, based on the reference, we evaluate the accuracy, completeness, and Chamfer distance of the estimated point clouds from VGGT, $\pi^3$, COLMAP MVS and 2DGS. 
To summarize, our contributions are as follows:
\begin{enumerate}[nosep]
    \item We introduce MVM-IOD, a novel industrial object dataset, to evaluate typical computer and machine vision tasks in an industrial setting. This addresses the lack of object-centric 3D reconstruction and pose estimation datasets focusing on industrial objects.
    \item We evaluate three camera pose estimation methods and four methods for image-based 3D reconstruction on MVM-IOD, to provide a baseline for future research and evaluations.
    \item We identify weaknesses of current state-of-the-art feed-forward methods in industrial applications and provide workarounds.
\end{enumerate}

\section{Related work}
\label{sec:related work}
Section \ref{sec:datasets} gives an overview of already existing datasets for image based 3D reconstruction. In Section \ref{sec:methods}, we take a look at different computer vision algorithms that solve tasks such as NVS, 3D reconstruction and camera pose estimation. 
\subsection{Datasets}
\label{sec:datasets}
 The DTU dataset \citep{aanaes2016large} contains a collection of 80 different scenes with 49 or 64 camera positions per scene. The capture setup consists of two cameras and a projector mounted to the end effector of a robot arm, generating images and structured-light reference scans from roughly the same viewing angles. This results in precise reference camera poses and 3D geometry. Most of the objects have a lot of texture, making it unfavorable for industrial applications, where objects often are textureless, reflective, transparent, or induce multiple of these challenges at the same time.
The Dataset of Industrial Metal Objects (DIMO) \citep{de2022dataset} contains textureless metal objects and uses different cameras mounted to the end effector of an industrial robot to capture the images. While DIMO contains a total of $42.620$ scenes, each scene is only captured from $13$ camera poses, making it less suited for NVS, as current NVS methods cannot unleash their full potential with such few images \citep{peng2024neurips}. Furthermore, the objects used in DIMO all have a very primitive geometry such as rectangular cuboids, limiting the relevance of the dataset for industrial applications.
A similar setup is used by \cite{durner2017experience} in the Top Hat Rail (THR) dataset, which captures images from $200$ camera poses per scene. However, there are only a total of four scenes, and the objects are all of a similar, rectangular shape, not being a hard challenge for image-based 3D reconstruction. The T-Less dataset \citep{hodan2017t} contains $ $ scenes with $504$ images each of multiple textureless objects placed on a turntable equipped with marker fields for camera pose estimation. Principally, T-Less is suited for evaluating 3D reconstruction and camera pose estimation methods of industrial objects, however, it lacks objects with complex geometry and reflective or transparent surfaces. 
\cite{hillemann2024novel} use the same capturing hardware as we, and a subset of the objects. The main difference is the sampling strategy of the camera poses: \cite{hillemann2024novel} sample the poses on a latitude/longitude grid, while we sample the camera poses equidistantly on a hemisphere, which avoids a high sampling density at the pole. Furthermore, we include additional 12 scenes, and evaluate the 3D reconstructions. \cite{langendorfer2025industry} uses the same camera pose sampling strategy and two of the objects used in MVM-IOD for evaluating the 2D and 3D reconstructions of NeRF \citep{mildenhall2021nerf} and Gaussian Splatting \citep{kerbl20233d} methods. In comparison, MVM-IOD contains 14 additional scenes, is evaluated on a broader set of methods, and is made available publicly. 
The NeRFBK \citep{yan2023nerfbk} dataset is tailored to the evaluation of NVS algorithms. It contains a collection of 21 indoor and outdoor scenes, though only a fraction of them is relevant in an industrial context. The images of the relevant scenes are captured by smartphone cameras or are rendered synthetically, that do not reflect industrial capture scenarios.
The NeRF synthetic dataset \citep{mildenhall2021nerf} consists of nine synthetic scenes with 600 images each. While not relevant for industrial applications, it can be beneficial to use synthetic scenes because of their perfect ground truth data.

With MVM-IOD, we aim to fill the outlined lack of benchmark datasets in the industrial domain for image based 3D reconstruction and camera pose estimation. MVM-IOD consists exclusively of industry-relevant objects with complex geometry, making it more applicable to industrial scenarios than DTU and THR. Additionally, in contrast to DIMO, it consists of densely sampled images which supports NVS. Furthermore, all scenes are captured with a realistic industrial setup, and do not contain synthetic data in comparison to NeRFBK.

\subsection{Methods}
\label{sec:methods}
\paragraph{SfM and MVS.} SfM describes the problem of finding the interior orientation and camera pose along with a sparse point cloud from a set of uncalibrated images \citep{agarwal2011building}. In contrast to current learning-based methods, traditional, rule-based methods divide this process in three stages: keypoint extraction, keypoint matching, and a bundle adjustment.
MVS describes the problem of extracting a dense point cloud from a set of images. Typically, the SfM result is used as an input along with the images \citep{furukawa2015multi,schonberger2016pixelwise}. COLMAP \citep{schoenberger2016sfm,schonberger2016pixelwise} is a widespread implementation of both SfM and MVS and the long-time standard.

\paragraph{NVS. }The task of generating novel views of a scene from a set of images with associated interior orientation and camera pose has seen a lot of attention in recent years. \cite{mildenhall2021nerf} introduced the concept of NeRFs that represent scenes as neural radiance fields, which allow for the generation of accurate novel views, with the drawback of long training times for each single scene. 3D Gaussian Splatting (3DGS, \cite{kerbl20233d}) uses Gaussian primitives with view-dependent color encoding to reconstruct a scene. This approach allows for faster training times per scene at a comparable quality of the generated novel views compared to NeRFs \citep{hillemann2024novel}. Subsequent works, such as \cite{guedon2024sugar}, \cite{chen2024pgsr}, and \cite{huang20242d} use the basic 3DGS concept, but improve the 3D reconstruction that can be extracted from the Gaussians. Specifically, \cite{huang20242d} do this by using 2D instead of 3D Gaussian primitives, which resolves multi-view geometry inconsistencies.

\paragraph{Feed-forward visual geometry reconstruction. }One of the main disadvantages of the mentioned SfM, MVS, and NVS algorithms is that they need to perform a training for each new scene. With the advent of new neural network structures \citep{vaswani2017attention} as well as abundant training data for visual tasks such as \cite{dai2017scannet, yao2020blendedmvs,reizenstein2021common}, new feed-forward visual geometry reconstruction methods arise. DUSt3R \citep{wang2024dust3r} uses a Visual Transformer architecture to regress pointmaps and camera poses from two images, essentially solving the previously mentioned traditional MVS task in a feed-forward manner by learning a generalizable, scene-agnostic representation. MASt3R \citep{leroy2024grounding} builds upon this by introducing local feature heads, making it robust against large viewpoint changes. Recently, VGGT \citep{wang2025vggt} used a transformer with global and local attention to infer pointmaps, depth maps, point tracks, and camera poses from an arbitrary number of input images. By simultaneously estimating these redundant properties, they manage to outperform methods like DUST3R and MAST3R. VGGT can generate a 3D reconstruction directly from the point maps, as well as by projecting the pixels of the depth maps via the camera poses and the interior orientation. $\pi^3$ \citep{wang2025pi} builds upon the idea of VGGT, but it removes the inductive bias of a fixed reference view and adds a local smoothness. The inference time of these methods is in the range of a few seconds, depending on the number of images. Furthermore, they are reporting a competitive quality compared to the results obtained by SfM and MVS. However, almost none of the testing data that was used up to now for evaluating VGGT and $\pi^3$ contains industry-relevant scenes or objects, making an evaluation in such critical applications highly relevant.

\section{The MVM-IOD}
\label{sec:data}
\begin{figure}[h]
  \centering
  \subfloat[Object 1]{\includegraphics[width=0.15\textwidth]{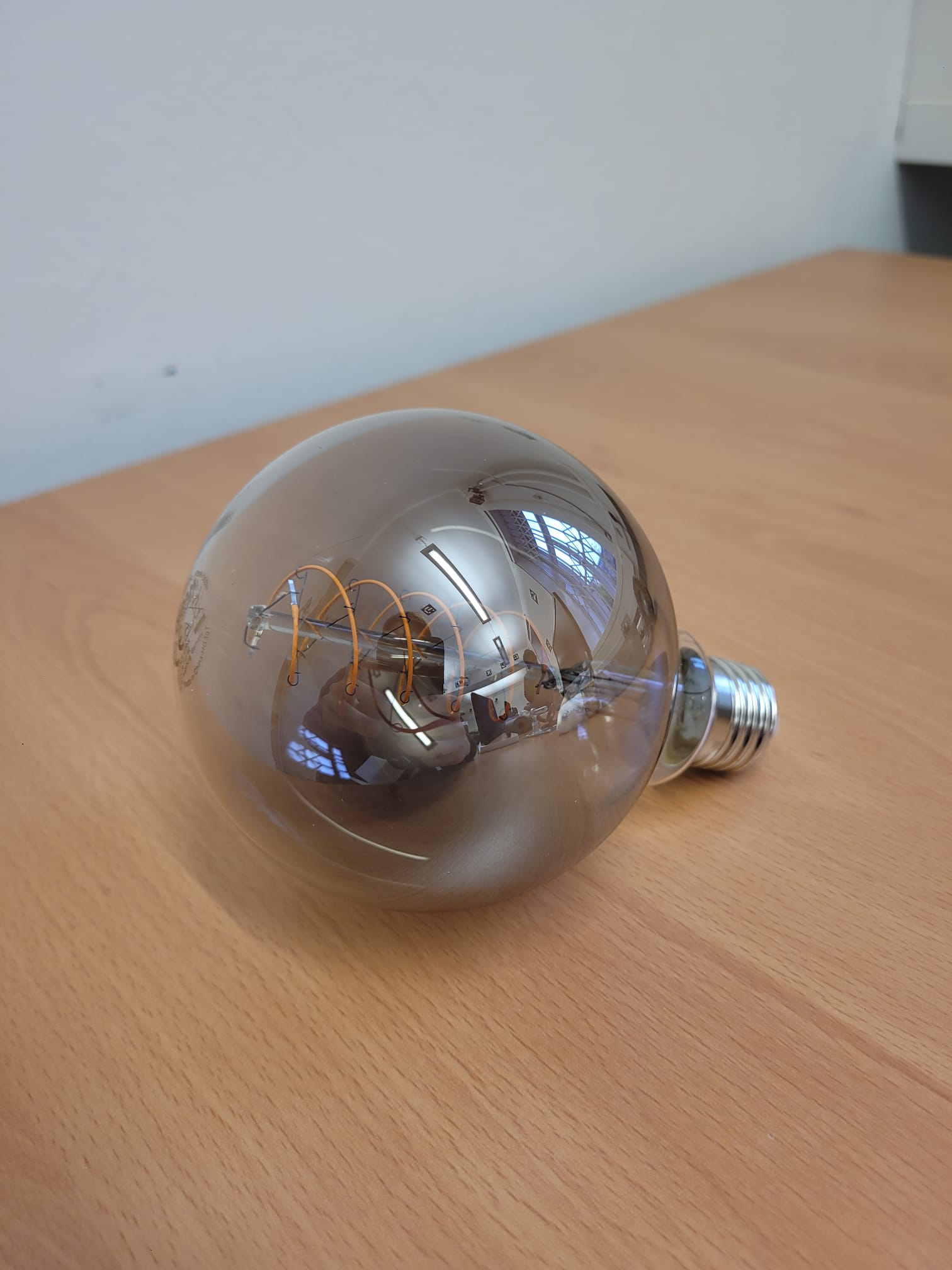}}
  \hfill
  \subfloat[Object 2]{\includegraphics[width=0.15\textwidth]{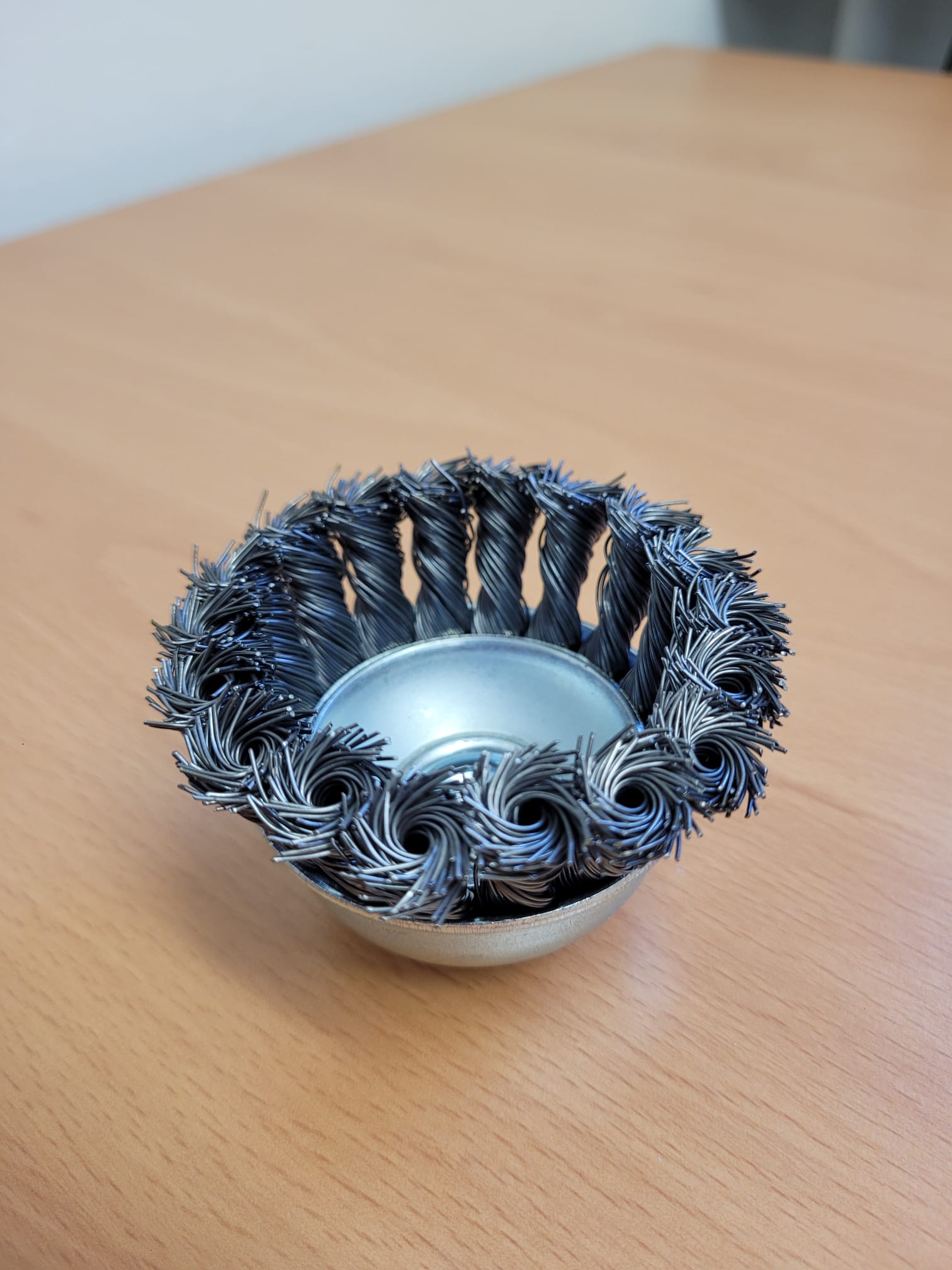}}
  \hfill
  \subfloat[Object 3]{\includegraphics[width=0.15\textwidth]{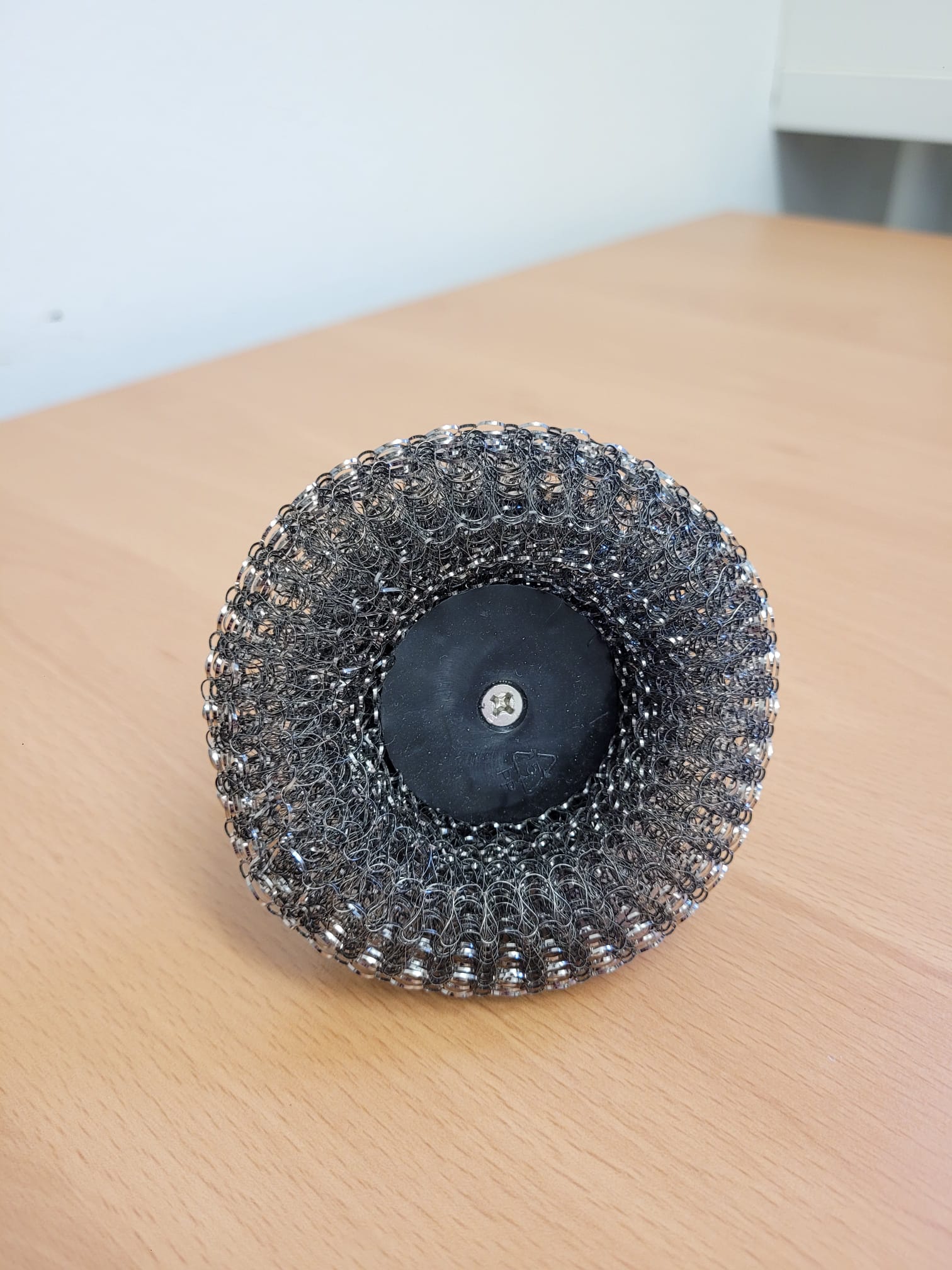}}
  \hfill
  \subfloat[Object 4]{\includegraphics[width=0.15\textwidth]{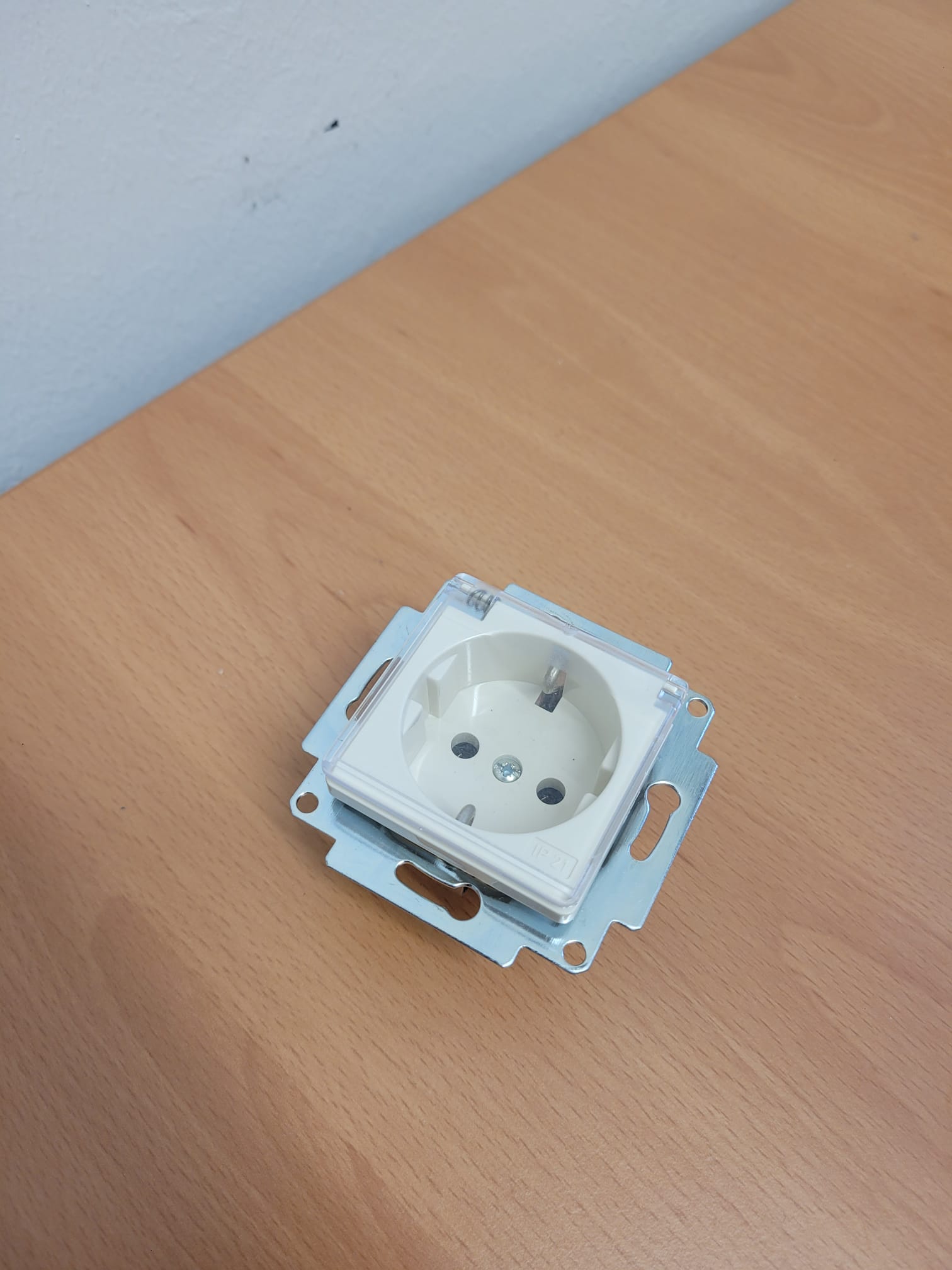}}\hfill
  \subfloat[Object 5]{\includegraphics[width=0.15\textwidth]{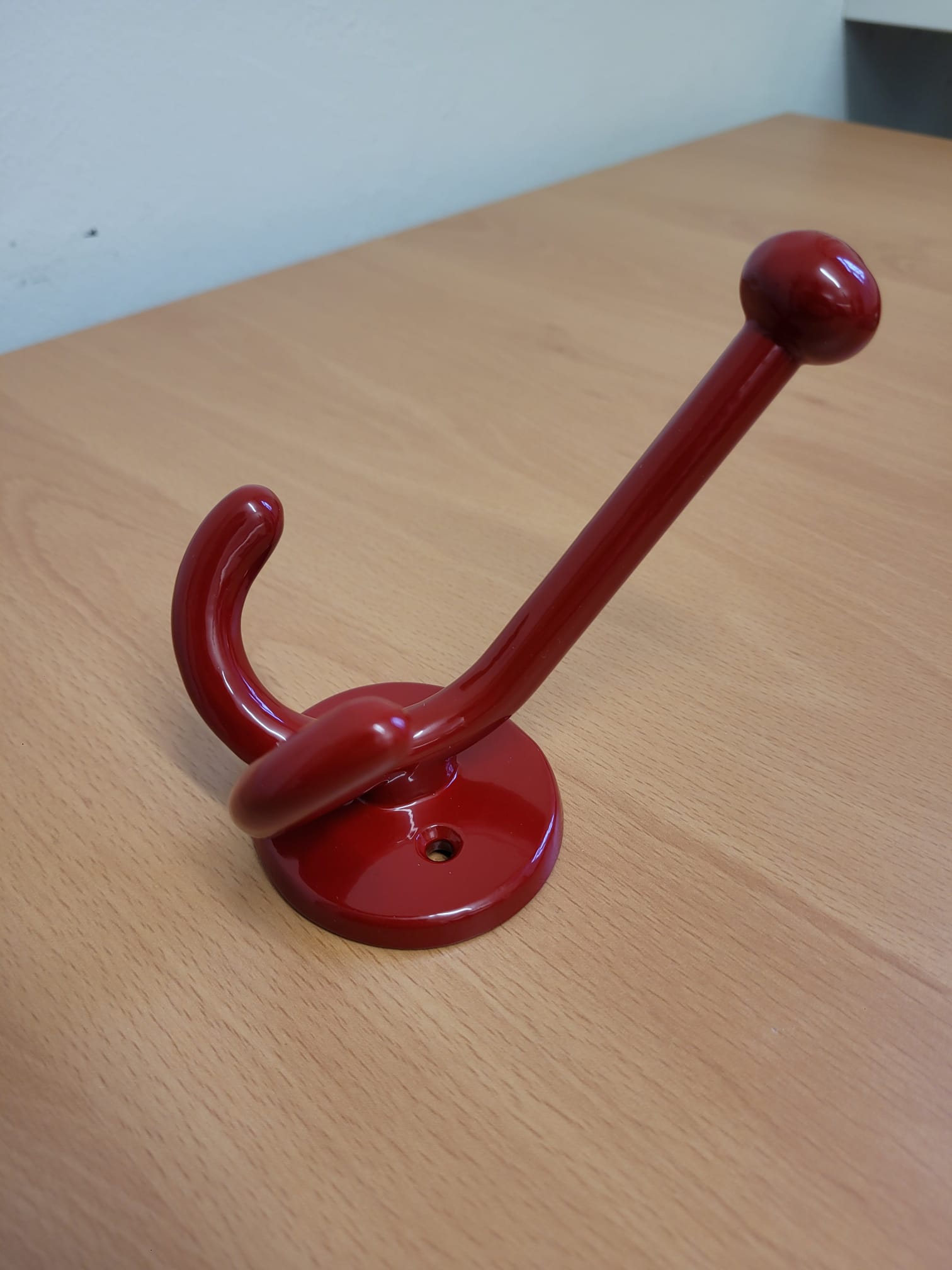}}\hfill
  \subfloat[Object 6]{\includegraphics[width=0.15\textwidth]{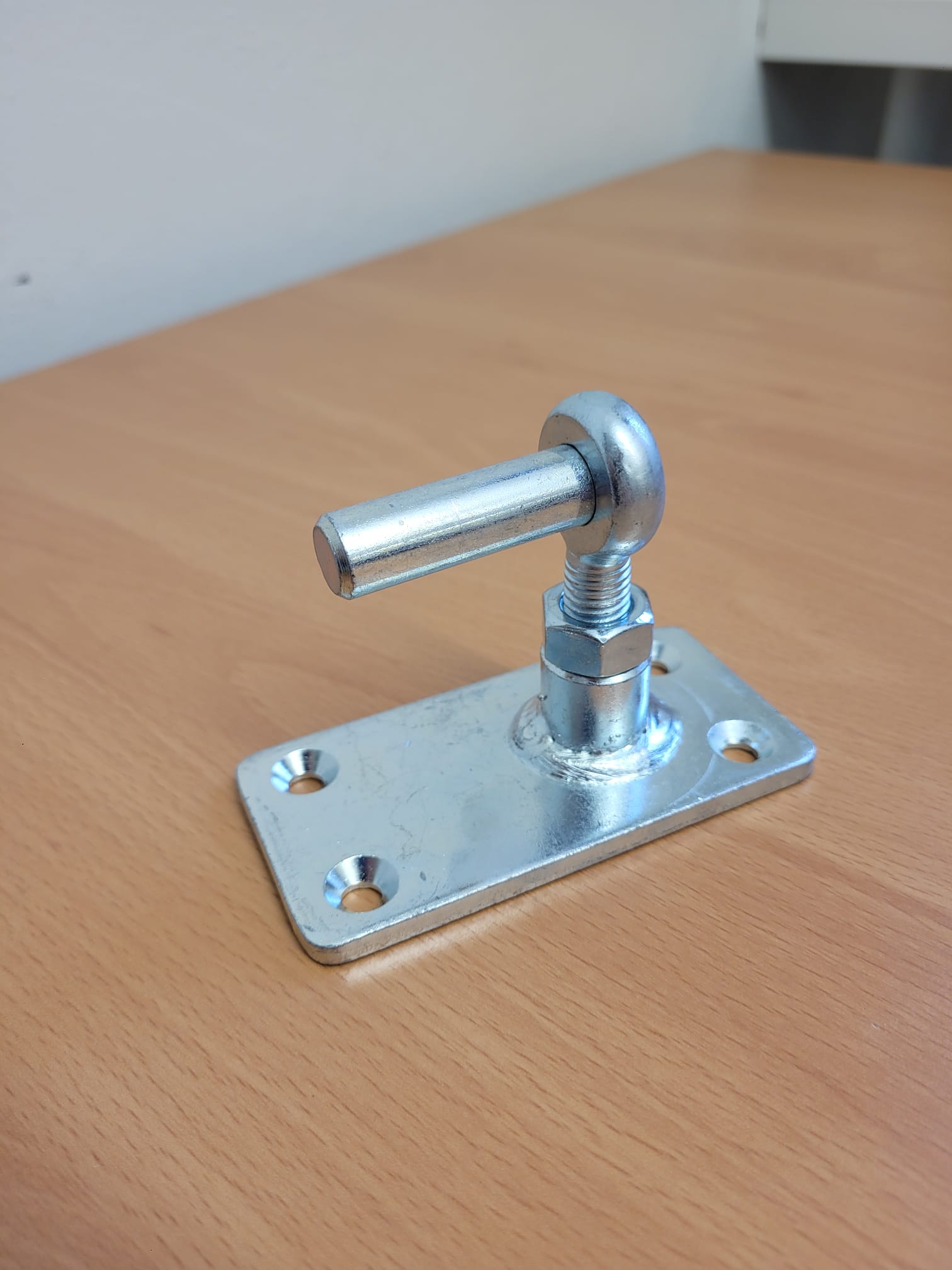}}\hfill
  
  \subfloat[Object 7]{\includegraphics[width=0.15\textwidth]{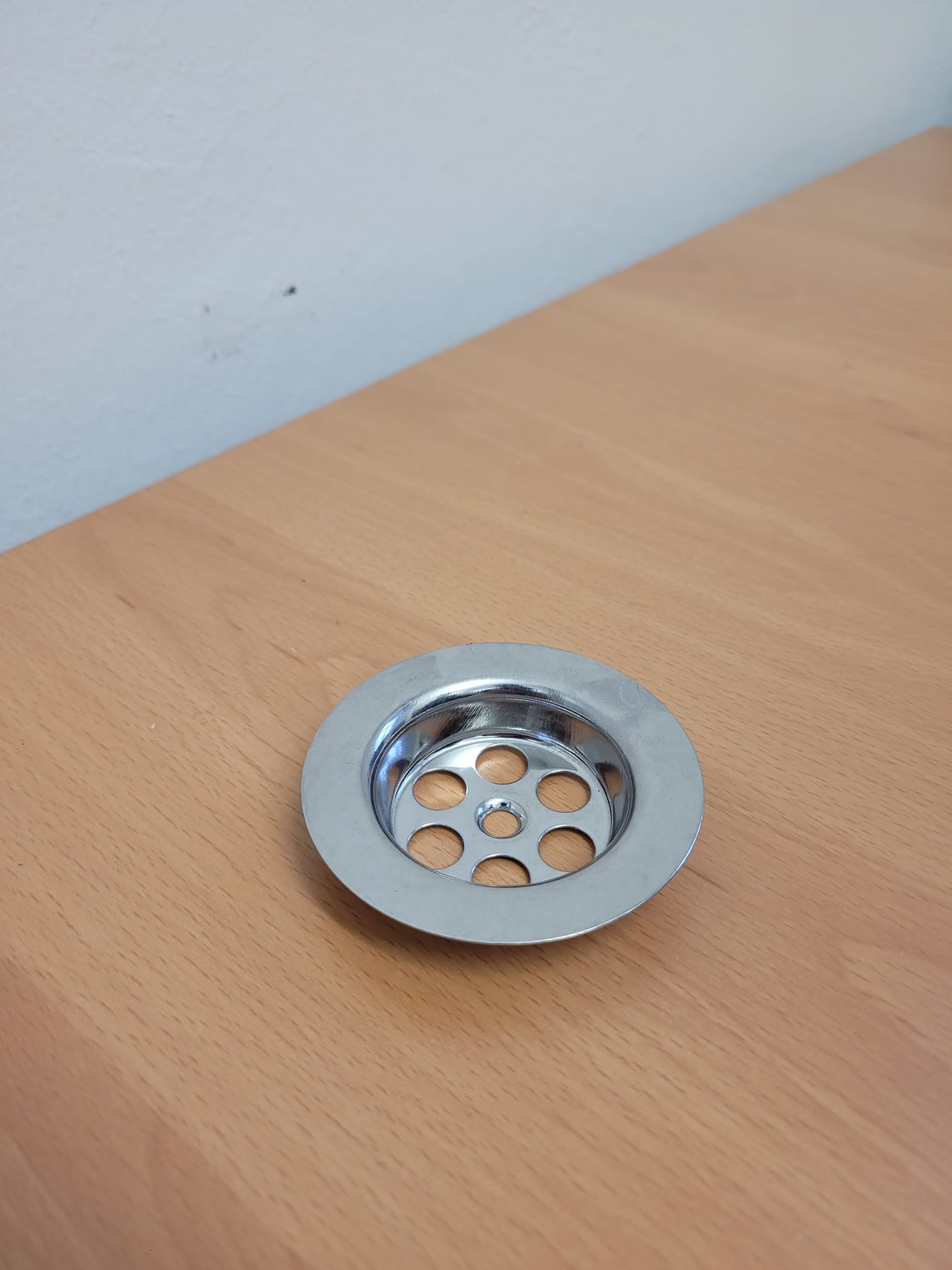}}\hfill
  \subfloat[Object 8]{\includegraphics[width=0.15\textwidth]{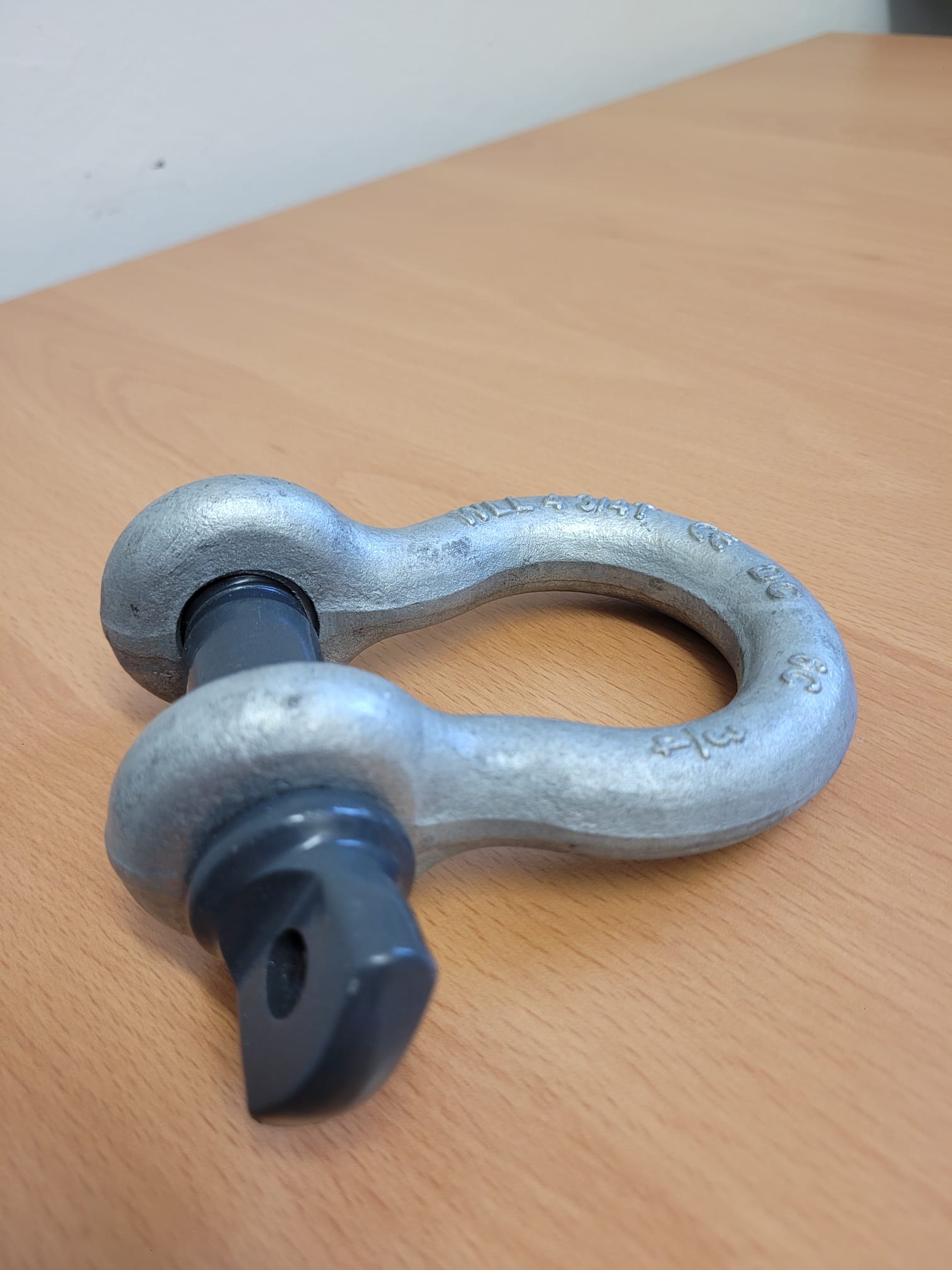}}\hfill
  \subfloat[Object 9]{\includegraphics[width=0.15\textwidth]{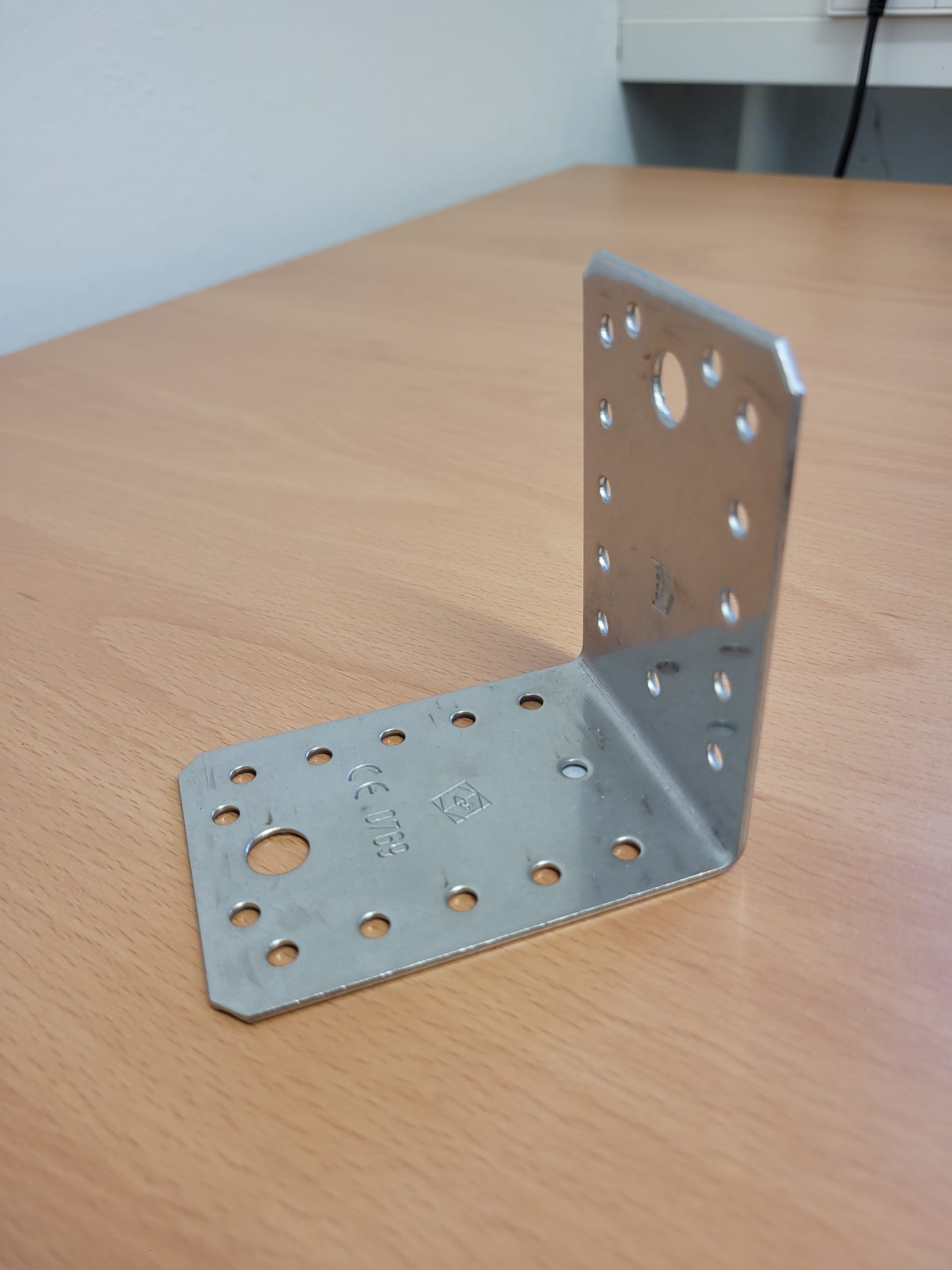}}\hfill

  \caption{Objects contained in the dataset, Figure adapted from \cite{langendorfer2025industry}}
  \label{fig:objects_img}
\end{figure}

In this section, we outline how the MVM-IOD is generated. First, detailed descriptions of the captured objects are given. Subsequently, we give details on the capture setup and the used hardware, and finally we describe how we obtain the reference data for the point clouds and the camera poses.
\paragraph{Objects. }The dataset contains industry-relevant objects, as seen in Figure \ref{fig:objects_img}, with sizes ranging from 7 to 15\Unit{cm} diameter. The objects cover various properties, such as transparency, reflectance, lack of texture, and complex geometry, which are challenging for vision-based reconstruction algorithms. The focus of MVM-IOD is to represent typical industrial applications. Each object is captured with two different background choices. The first background is an aluminum table with profile rays, which provides a rich texture for keypoint extraction and matching. The second background is a homogeneous green background, to mimic poor texture in the background, which also often occurs in industrial applications. The background choices are referred to as alu and green, respectively.

\paragraph{Hardware and Setup. }All images are acquired by an IDS U3-31J0CP-C-HQ Rev 2.2 RGB industrial camera with a Tamron M23FM08 $8$\,mm lens. The captured images have an image resolution of $2840\times2840$\Unit{px}. The camera is mounted on the end effector of a Universal Robots UR3e robot arm with a pose repeatability error of $\pm0.03$\Unit{mm} according to ISO 9283. The robot was calibrated by using the approach of \cite{ulrich2024robotcal}. It is important to note that the camera is mounted so that all images appear upside-down due to space restrictions. A camera calibration as well as an uncertainty-aware hand--eye calibration \citep{ulrich2021uncertainty,ulrich2024uncertainty} is performed with a total of 25 calibration images. We define a pose $\TMatrix{b}{H}{a}\in \mathrm{SE(3)}$, indicating a $4\times4$ rigid transformation from coordinate system $a$ to coordinate system $b$. The relevant coordinate systems in our setup are base ($b$), world ($w$), tool ($t$) and camera ($c$). 

 The images of each scene are captured at the same  $\TMatrix{w}{H}{c}$ camera poses with the object in the center of the scene. With the result of the hand--eye calibration, we are able to calculate  the corresponding $\TMatrix{t}{H}{b}$ by matrix multiplication, which is passed to the robot controller to move the robot arm. The $\TMatrix{w}{H}{c}$ poses are sampled equidistantly on a sphere with radius of $20$\Unit{cm} via the algorithm described by \cite{gonzalez2010measurement} up to a maximum zenith angle of $35$\Unit{$^\circ$}, resulting in a total of 300 $\TMatrix{w}{H}{c}$ poses per scene. The radius and the maximum zenith angle are physically limited by the robot workspace. 
We refer to the $\TMatrix{w}{H}{c}$ poses obtained in this way as the robot-derived camera poses, which can be used as a metric alternative to the COLMAP SfM camera pose estimation. We report the Root Mean Square Translation (RMST), the Root Mean Square Rotation (RMSR), and the Reprojection Root Mean Square Error (RRMSE) of the hand--eye calibration, as implemented by \cite{hillemann2024novel}: $\mathrm{RMST}=0.11$\Unit{mm}, $\mathrm{RMSR}=0.030$\Unit{$^\circ$}, and $\mathrm{RRMSE}=0.94$\Unit{px}. The mean standard deviation is $\sigma_{\mathrm{trans}}=0.07$\Unit{mm} for translation and $\sigma_{\mathrm{rot}}=0.018^\circ$ for rotation, which is roughly two times the robot pose repeatability error.

\paragraph{Reference. }The reference point cloud of the objects is captured with a Hexagon AS1 line scanner mounted to an AA85 measurement arm with a system accuracy of $0.047$\Unit{mm} according to ISO 10360-8 D. All reconstructed points in the reference point cloud that do not belong to the scanned object have been filtered out manually. Because the reference point cloud is not represented in the robot coordinate system, a registration is necessary. 

The reference for the $\TMatrix{w}{H}{c}$ poses is obtained by capturing a scene with a calibration plate in the center of the scene, which is acquired with the same 300 $\TMatrix{t}{H}{b}$ poses by a calibrated camera and performing spatial resectioning with fixed internal camera parameters. The reference robot poses are still influenced by the pose repeatability error of the robot in each dataset. We report a RRMSE of $0.39\,$px for the reference poses.

\section{Experiments}
\label{sec:Experiments}
All experiments are performed on a computer with an NVIDIA\,\textsuperscript{\tiny\textregistered} A100 with $40$\Unit{GB} of VRAM. We choose to compare modern feed-forward geometry reconstruction methods to assess their use for real-time industry applications. \\
\subsection{Evaluated Methods}
\label{sec:Methods}
\paragraph{COLMAP. }We use COLMAP V3.13.0 \citep{schoenberger2016sfm,schonberger2016pixelwise} to generate the COLMAP SfM and MVS results. COLMAP SfM takes roughly 5 to 10 minutes, depending on the object and the background choice. COLMAP MVS takes about 3 to 4 hours.

\paragraph{VGGT. }We use the pretrained VGGT-1B checkpoint with $ 1.26$\Unit{B}  parameters for training. All images are resized by default to a size of $518\times518$\Unit{px}. By default, non-square images are zero-padded. We report an inference time of 17 seconds. VGGT can estimate point clouds via directly via the point maps, which we refer to as VGGT point, or by projecting the estimated depth maps with the estimated camera pose and interior orientation, which we refer to as VGGT depth. 

\paragraph{\bm{$\pi^3$}. }All square images are resized to $504\times504$\Unit{px}, while non-square images get resized to $588\times420$\Unit{px}. We use the standard pretrained model with $0.96$\Unit{B} parameters for inference. We report an inference time of 17 seconds.

\paragraph{2DGS. }For 2DGS input, we use the COLMAP SfM result to undistort the input images, as well as for the camera poses and the initial point cloud. We use the default training parameters and train for $30000$ iterations, which roughly takes 3 hours.

The tested methods use different amounts of VRAM. VGGT uses too many resources to process all $300$ images of each dataset at once. Therefore, we subsample each dataset randomly to $n=100$ images, but keep the same random subsampling for all datasets and experiments. For COLMAP and 2DGS, all images are processed in full resolution.

\subsection{Evaluation}
\label{sec:Evaluation}
\begin{figure}[bp!]
    \centering
  \subfloat[Original image]{\includegraphics[width=0.2\textwidth]{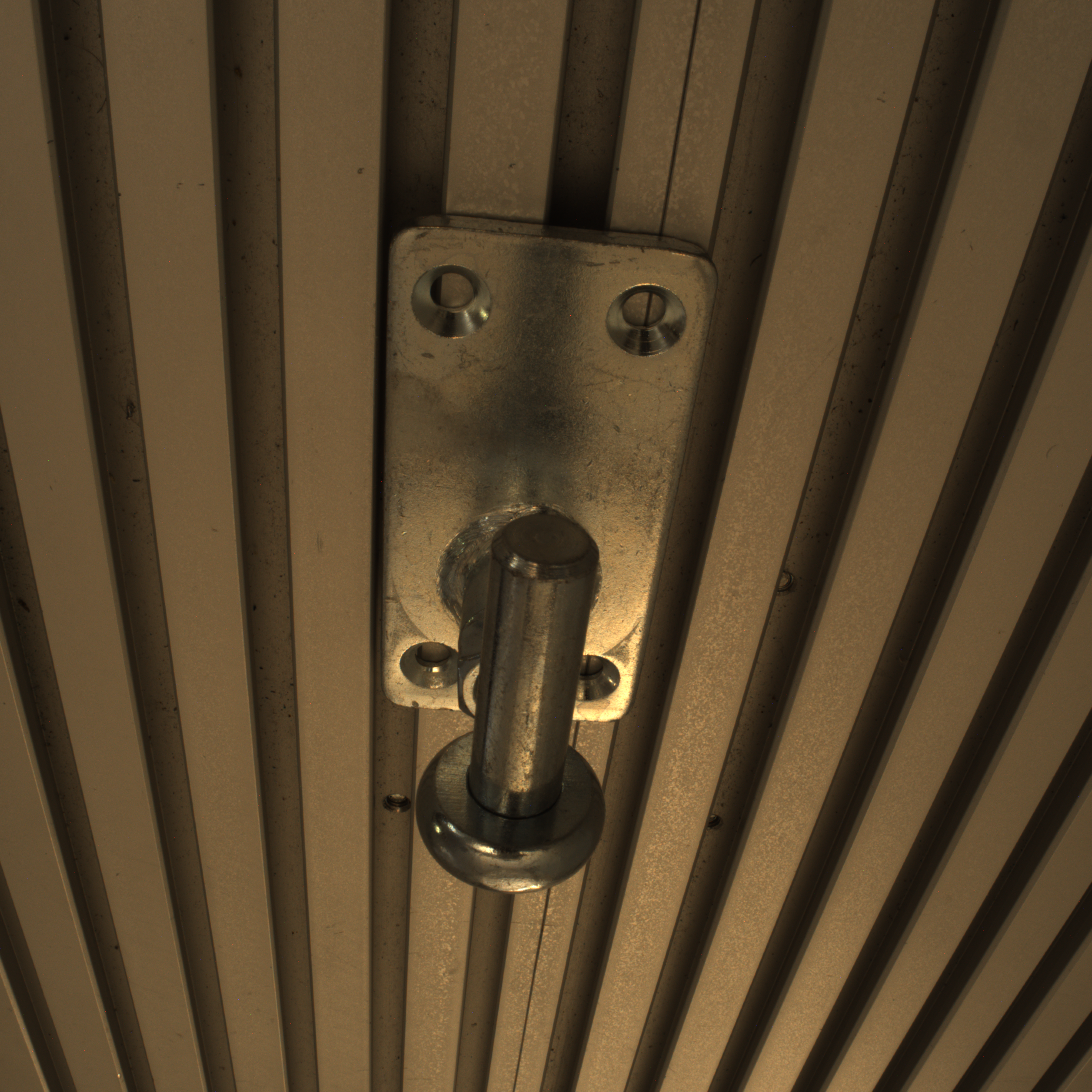}}
  \hspace{1em}
  \subfloat[Preprocessed image]{\includegraphics[width=0.2\textwidth]{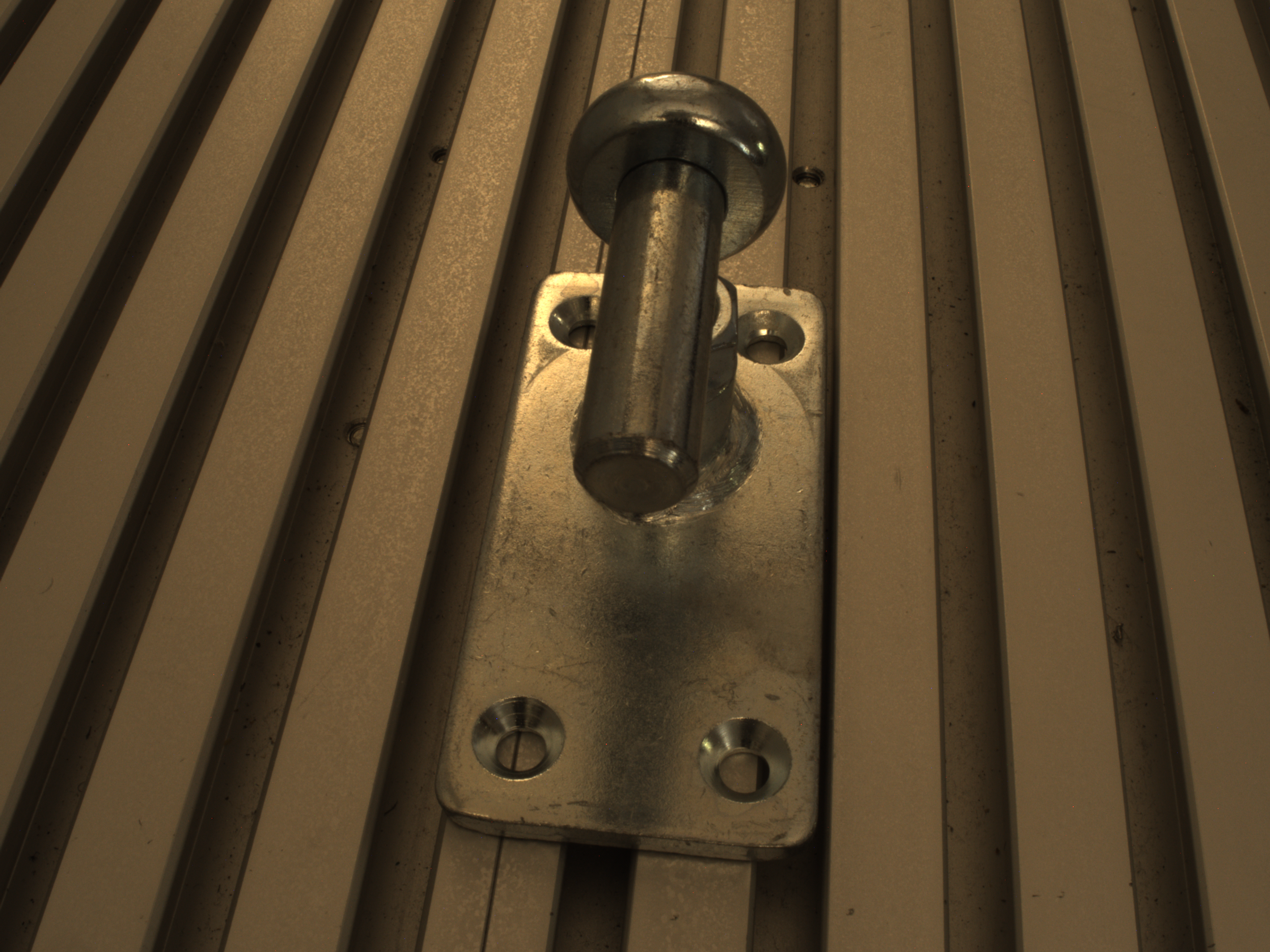}}\\
  \subfloat[VGGT depth with original images]{\includegraphics[width=0.25\textwidth]{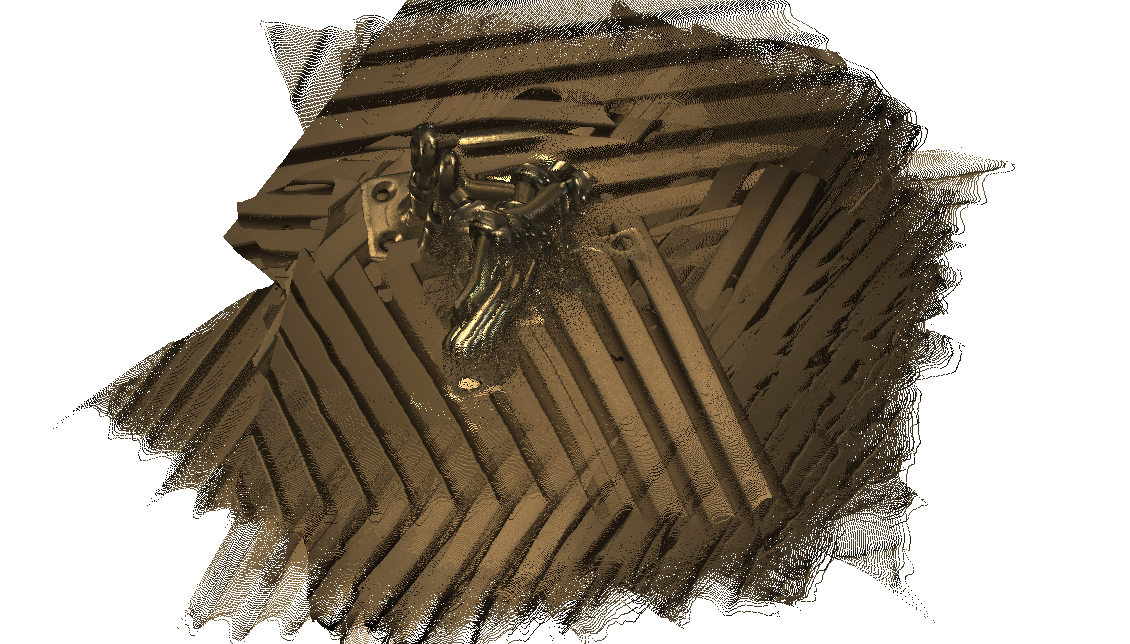}}
    \subfloat[VGGT depth with preprocessed images]{\includegraphics[width=0.25\textwidth]{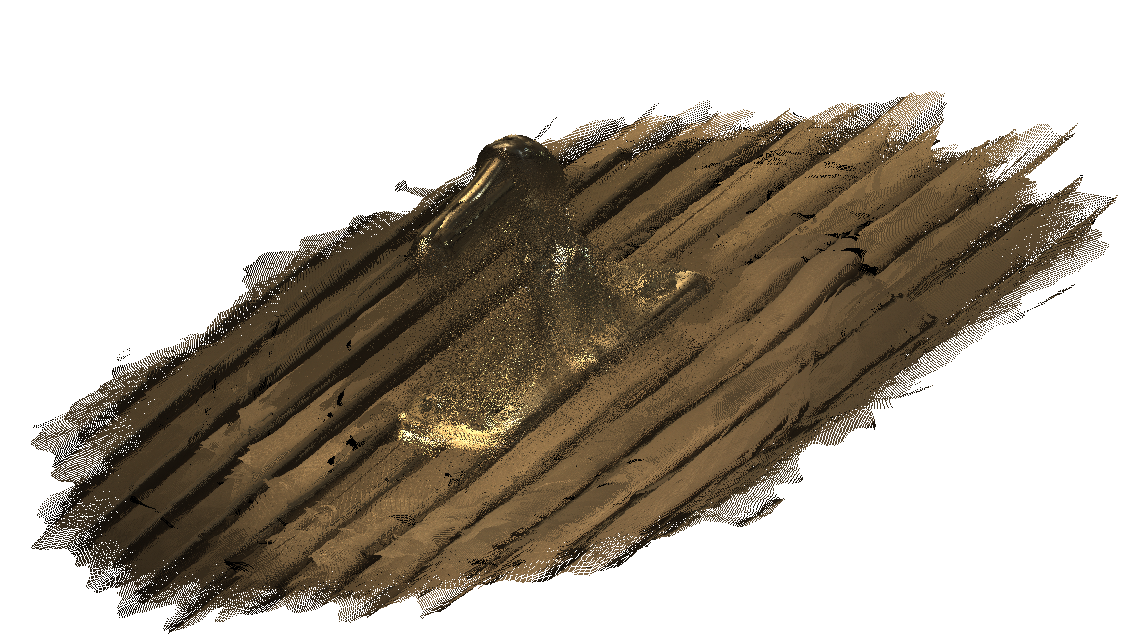}}\\
      \subfloat[VGGT point with original images]{\includegraphics[width=0.25\textwidth]{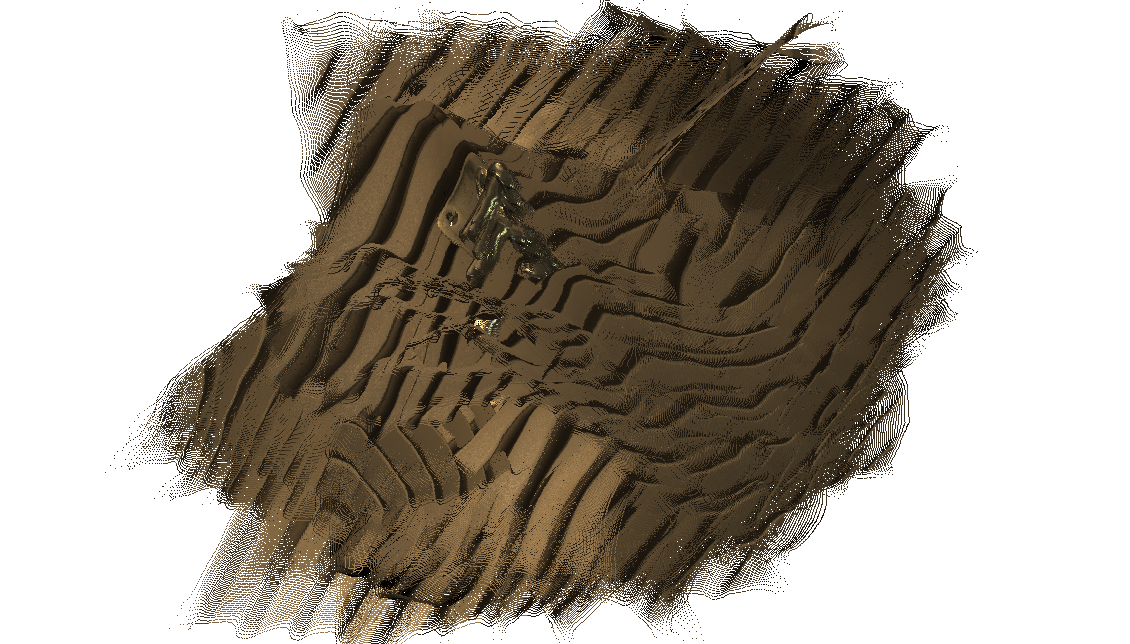}}
    \subfloat[VGGT point with preprocessed images]{\includegraphics[width=0.25\textwidth]{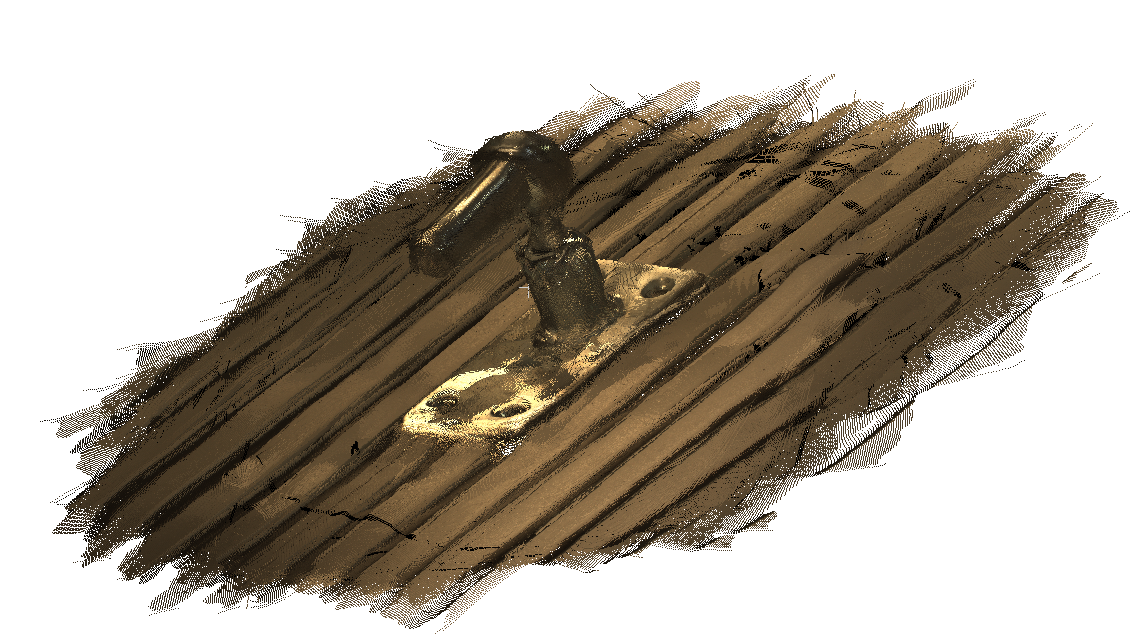}}\\
    \subfloat[$\pi^3$ with original images]{\includegraphics[width=0.25\textwidth]{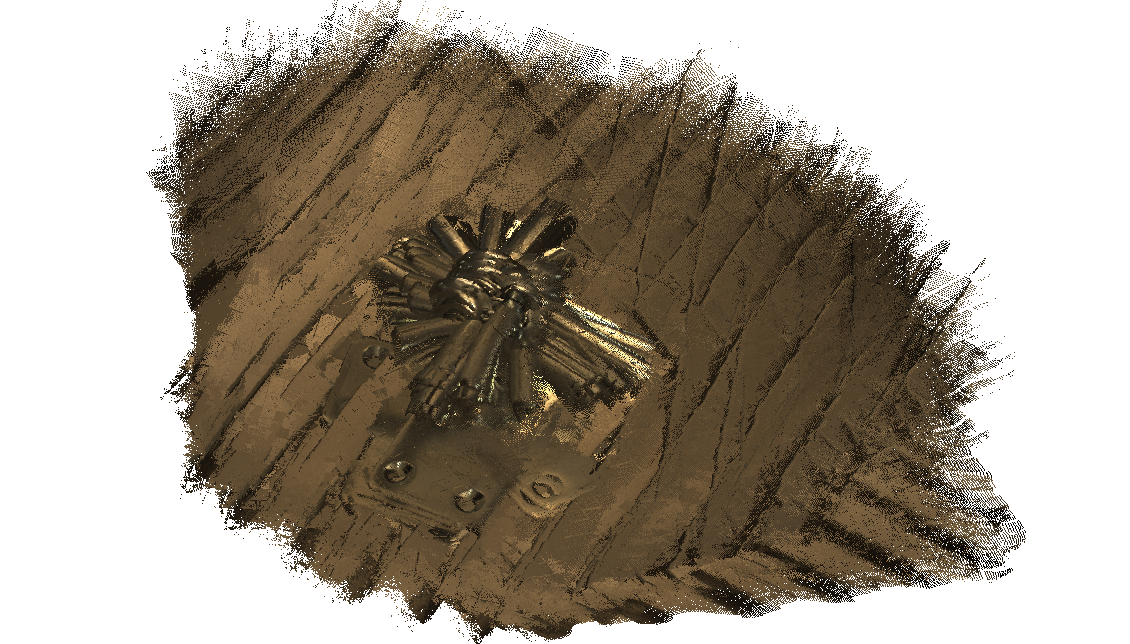}}
    \subfloat[$\pi^3$ with preprocessed images]{\includegraphics[width=0.25\textwidth]{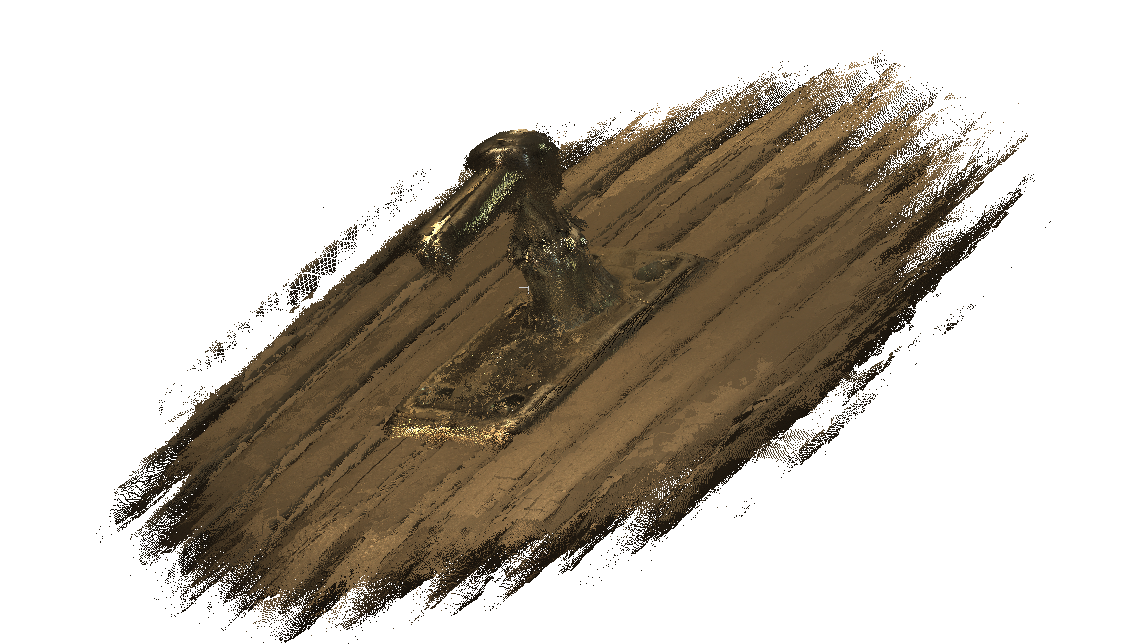}}
    
    \caption{a) and b) show images of object 6 with  background alu before (original) and after preprocessing by rotating the image by $180$\Unit{$^\circ$} and changing the aspect ratio to $4:3$. c), e), and g) show the VGGT point, VGGT depth and $\pi^3$ point clouds when using the original images. d), f), and h) show the same point clouds of VGGT and $\pi^3$ when using the preprocessed images.}
    \label{fig:point_cloud_fail}
\end{figure}

First, we show that VGGT and $\pi^3$ neither generate accurate camera poses nor accurate 3D reconstruction with the original MVM-IOD images and how to improve the results. Then, we evaluate both the accuracy of the poses and the accuracy and completeness of the generated point cloud of the different methods by comparing the results with the reference. A comparison of the internal camera parameters is not performed, as VGGT and $\pi^3$ do not estimate the distortion parameters. However, methods that estimate the distortion parameters can also be assessed with MVM-IOD.
\paragraph{Image preprocessing for VGGT and $\pi^3$.}
\label{sec:Image preprocessing}One initial insight is that the images of the dataset are difficult to process with feed-forward geometry transformers such as VGGT and $\pi^3$, and the 3D reconstruction as well as the pose estimation is visibly off most of the time with the original images, which can be seen, for example, in Figure \ref{fig:point_cloud_fail} for object 6 with background alu. This makes a registration and subsequent comparison to the reference point cloud nearly impossible because the reconstructed object is broken into multiple instances with different orientations. The most likely reason for this is that the image contents and the image format are out of domain for these models, which were not trained on similar data.
 It is shown empirically that by rotating the images by $180$\Unit{$^\circ$} and changing the aspect ratio from $1:1$ to $4:3$, VGGT and $\pi^3$ are able to drastically improve the pose estimation and 3D reconstruction accuracy of the scenes in most cases (Table \ref{tab:pose_acc_combined}). These preprocessing steps are motivated by the fact that the camera is mounted upside down to the end-effector, obviously making the scenes appear unusual for the trained models. Also, most of the training images used for VGGT and $\pi^3$ are non-square. Therefore, the MVM-IOD images are adapted to be more in-distribution also in this regard. Applying only one of the two preprocessing steps did not achieve satisfying improvements.\\

\paragraph{Camera Pose Evaluation.}
\label{sec:Pose Comparison}
For the evaluation of the camera poses, the EVO tool \citep{grupp2017evo} is used. Each set of estimated camera poses is compared to the corresponding reference poses obtained as described in Section \ref{sec:data}. As COLMAP SfM, VGGT, and $\pi^3$ do not output metric camera poses or point clouds, an alignment and scaling to the reference poses is necessary to enable a comparison. This is achieved by applying the Umeyama algorithm \citep{umeyama1991least}. Then, the Absolute Pose Error $\Matrix{H}_{\mathrm{APE},i}$ is calculated as
\begin{equation}
\Matrix{H}_{\mathrm{APE},i}=\Matrix{H}_{\mathrm{est},i}^{-1}\Matrix{H}_{\mathrm{ref},i}\in\mathrm{SE(3)} \enspace .
    \label{eq:APE}
\end{equation}
Where $\Matrix{H}_{\mathrm{est},i}$ is the estimated camera pose and $\Matrix{H}_{\mathrm{ref},i}$ the reference pose. The $\Matrix{H}_{\mathrm{APE},i}$ between a set of $N$ reference and estimated poses is used to calculate the Mean Translation Error (MTE, Equation~(\ref{eq:MTE})) and the Mean Rotation Error (MRE, Equation~(\ref{eq:MRE})):
\begin{equation}
    \mathrm{MTE}=\frac{1}{N}\sum_{i=1}^N \lvert \lvert\mathrm{trans}(\Matrix{H}_{\mathrm{APE},i})\rvert\rvert
    \label{eq:MTE}
\end{equation}
\begin{equation}
    \mathrm{MRE}=\frac{1}{N}\sum_{i=1}^N \lvert \mathrm{angle}(\mathrm{log_{SO(3)}}(\mathrm{rot}(\Matrix{H}_{\mathrm{APE},i})))\rvert
    \label{eq:MRE}
\end{equation}
Here, $\mathrm{trans}(\cdot)$ and $\mathrm{rot}(\cdot)$ are the translation and rotation parts of the $\Matrix{H}_{\mathrm{APE},i}$ and $\mathrm{angle}(\mathrm{log_{SO(3)}(\cdot)})$ is the Rodrigues angle in degrees.
Moreover, the RRMSE (Section \ref{sec:data}) of the COLMAP SfM and robot-derived camera poses is reported. Of course, the RRMSE of the robot-derived camera poses is constant over all objects in the dataset at $1.05$\Unit{px}, while the mean RRMSE over all COLMAP SfM results is $1.17$\Unit{px} with $\sigma_{\mathrm{RRMSE}}=0.096$\Unit{px}. RRMSE values for VGGT and $\pi^3$ are not calculated, because both methods reconstruct a point for each pixel and hence RRMSE would be zero. While it would be possible to perform a bundle adjustment after inference to refine the results, this approach would not evaluate the performance of the actual methods and would compromise their real-time capability. The resulting MTE and MRE are reported in Table \ref{tab:pose_acc_combined} for all methods and all scenes in the dataset.\begin{table*}[h!]
    \centering    \caption{
    Mean Translation Error (MTE) and Mean Rotation Error (MRE) for all dataset objects with both background choices (green/alu). For VGGT and $\pi^3$, we use the preprocessed images as described in Section \ref{sec:Image preprocessing}, while we use the original images for COLMAP SfM and robot-derived. Still, we report the metrics of VGGT and $\pi^3$ using the original images in parentheses and mark them in red if they are worse and green otherwise. Lowest metric for each object and background combination is print bold. }

    \small
    \begin{adjustbox}{width=\textwidth,center}

    \begin{tabular}{lcccccccccccc}
        \textbf{Method} & \textbf{Background} & \textbf{Object 1} & \textbf{Object 2} & \textbf{Object 3} & \textbf{Object 4} & \textbf{Object 5} & \textbf{Object 6} & \textbf{Object 7} & \textbf{Object 8} & \textbf{Object 9}&\textbf{Mean} \\
        \midrule
        \multicolumn{12}{c}{\textit{Mean Translation Error (MTE) [mm] $\downarrow$}} \\
        \midrule
        Robot-derived &green/alu & \textbf{0.10} & \textbf{0.10} & \textbf{0.10} & \textbf{0.10} & \textbf{0.10} & \textbf{0.10} & \textbf{0.10} & \textbf{0.10}& \textbf{0.10} &\textbf{0.10}\\
        \midrule
        $\pi^3$ & green & 5.34 (\textcolor{green!60!black}{4.65}) & 3.88 (\textcolor{red}{40.88}) & 5.75 (\textcolor{red}{22.47}) & 4.94 (\textcolor{red}{62.98}) & 4.56 (\textcolor{red}{5.62}) & 3.49 (\textcolor{red}{5.12}) & 14.10 (\textcolor{red}{65.42}) & 1.54 (\textcolor{red}{2.46}) & 3.04 (\textcolor{red}{10.39})& 5.18 (\textcolor{red}{24.44}) \\
        VGGT & green & 2.35 (\textcolor{red}{6.90}) & 4.61 (\textcolor{red}{70.96}) & 7.82 (\textcolor{red}{66.20}) & 3.07 (\textcolor{red}{57.68}) & 2.36 (\textcolor{red}{37.75}) & 2.10 (\textcolor{red}{44.01}) & 2.47 (\textcolor{red}{76.31}) & 2.65 (\textcolor{red}{4.47}) & 2.06 (\textcolor{red}{11.38}) &3.28 (\textcolor{red}{41.74})\\
        COLMAP SfM & green & - & - & - & - & - & \textbf{0.09} & - & 0.15 & 0.11 &0.12\\
        \midrule
        $\pi^3$ & alu & 6.42 (\textcolor{red}{60.62}) & 79.90 (\textcolor{green!60!black}{63.86}) & 79.70 (\textcolor{green!60!black}{58.25}) & 78.17 (\textcolor{green!60!black}{68.81}) & 4.22 (\textcolor{red}{52.66}) & 4.39 (\textcolor{red}{74.38}) & 72.11 (\textcolor{green!60!black}{39.34}) & 2.33 (\textcolor{red}{33.56}) & 2.86 (\textcolor{red}{69.15})& 36.70 (\textcolor{red}{57.84})\\
        VGGT & alu & 4.73 (\textcolor{red}{78.19}) & 80.21 (\textcolor{green!60!black}{79.95}) & 79.00 (\textcolor{green!60!black}{76.06}) & 79.11 (\textcolor{green!60!black}{77.38}) & 3.78 (\textcolor{red}{64.74}) & 4.25 (\textcolor{red}{78.05}) & 78.57 (\textcolor{red}{79.55}) & 3.20 (\textcolor{red}{50.29}) & 5.40 (\textcolor{red}{67.35})&37.58 (\textcolor{red}{72.40}) \\
        COLMAP SfM& alu & 0.27 & \textbf{0.07} & 0.17 & 0.22 & 0.24 & 0.14 & 0.26 & \textbf{0.08} & 0.24& 0.19\\
        \midrule
        \multicolumn{12}{c}{\textit{Mean Rotation Error (MRE) [\textdegree] $\downarrow$}} \\
        \midrule
        Robot-derived & green/alu & \textbf{0.08} & \textbf{0.08} & \textbf{0.08} & \textbf{0.08} & \textbf{0.08} & \textbf{0.08} & \textbf{0.08} & \textbf{0.08} & \textbf{0.08} &\textbf{0.08}\\
        \midrule
        $\pi^3$& green & 2.98 (\textcolor{red}{3.27}) & 2.39 (\textcolor{red}{29.27}) & 1.15 (\textcolor{red}{18.94}) & 2.17 (\textcolor{red}{43.06}) & 4.22 (\textcolor{green!60!black}{3.03}) & 1.16 (\textcolor{red}{2.29}) & 4.94 (\textcolor{red}{61.31}) & 1.32 (\textcolor{green!60!black}{1.16}) & 1.21 (\textcolor{red}{6.03})& 2.39 (\textcolor{red}{18.71})\\
        VGGT & green & 2.14 (\textcolor{red}{3.44}) & 3.64 (\textcolor{red}{170.10}) & 6.35 (\textcolor{red}{164.35}) & 1.12 (\textcolor{red}{40.49}) & 1.93 (\textcolor{red}{20.75}) & 1.71 (\textcolor{red}{15.12}) & 2.33 (\textcolor{red}{82.21}) & 1.78 (\textcolor{red}{4.01}) & 1.18 (\textcolor{red}{4.47}) &2.46 (\textcolor{red}{56.10})\\
        COLMAP SfM& green & - & - & - & - & - & 0.72 & - & 0.73 & 0.72& 0.72\\
        \midrule
        $\pi^3$ & alu & 6.68 (\textcolor{red}{66.28}) & 150.87 (\textcolor{red}{159.00}) & 98.44 (\textcolor{green!60!black}{54.01}) & 85.36 (\textcolor{red}{92.43}) & 1.82 (\textcolor{red}{162.58}) & 2.40 (\textcolor{red}{138.00}) & 79.44 (\textcolor{green!60!black}{24.28}) & 0.90 (\textcolor{red}{23.67}) & 2.29 (\textcolor{red}{76.91}) & 47.58 (\textcolor{red}{88.57})\\
        VGGT & alu & 2.06 (\textcolor{red}{99.49}) & 122.42 (\textcolor{green!60!black}{121.30}) & 124.30 (\textcolor{green!60!black}{95.12}) & 136.69 (\textcolor{green!60!black}{97.48}) & 2.04 (\textcolor{red}{55.48}) & 1.36 (\textcolor{red}{101.39}) & 135.00 (\textcolor{green!60!black}{113.64}) & 1.34 (\textcolor{red}{25.54}) & 1.91 (\textcolor{red}{143.70}) &58.57 (\textcolor{red}{94.79})\\
        COLMAP SfM& alu & 0.70 & 0.72 & 0.71 & 0.69 & 0.69 & 0.71 & 0.71 & 0.72 & 0.70 &0.71\\
                \midrule

    \end{tabular}
        \end{adjustbox}
    \label{tab:pose_acc_combined}

\end{table*} For VGGT and $\pi^3$, we report the results for the original images as well as for the preprocessed images as described in Section \ref{sec:Image preprocessing}. We observe that the COLMAP SfM and robot-derived MTE are in the same order of magnitude, though for some scenes with green background, COLMAP SfM completely fails likely due to missing texture. The robot-derived MRE is one order of magnitude lower than the COLMAP SfM MRE. In contrast to \cite{wang2025pi,wang2025vggt} our results show that both COLMAP SfM and robot-derived have significantly lower MRE and MTE values than VGGT and $\pi^3$ in all cases. For VGGT, preprocessing the images results in significantly lower MRE and MTE values in 14 cases while MRE and MTE stay roughly the same for objects 2, 3, 4, and 7 with the alu background. Roughly the same applies to $\pi^3$, though the MRE and MTE values increase for object 7 with background alu when applying the preprocessing to the images. Generally, it can be observed that VGGT and $\pi^3$ seemingly struggle with the alu background, as after preprocessing only scenes with background alu have large MRE and MTE values. At the same time, even without the preprocessed images, somewhat low MRE and MTE values are achieved for objects 1, 5/6 (only $\pi^3$) and 8 with green background.

\paragraph{3D Reconstruction.}
\label{sec:3D reconstruction}
We report the mean of accuracy and the mean of completeness, as well as the overall metric of the point clouds. The accuracy ($\mathrm{ACC_i}$) for each point $\Point{x}_i$ in the estimated point cloud $\Matrix{P}_{\mathrm{est}}$ is calculated as seen in Equation~(\ref{eq:acc}), while the completeness ($\mathrm{COMP_j}$) of each point $\Point{x}_j$ in the reference point cloud $\Matrix{P}_{\mathrm{ref}}$ is calculated as in Equation~(\ref{eq:comp}).
\begin{equation}
\mathrm{ACC}_i=\lvert\lvert \Point{x}_i-\mathrm{NN}(\Point{x}_i,\Matrix{P}_{\mathrm{ref}})\rvert\rvert_2
    \label{eq:acc}
\end{equation}
\begin{equation}
\mathrm{COMP}_j=\lvert\lvert \Point{x}_j-\mathrm{NN}(\Point{x}_j,\Matrix{P}_{\mathrm{est}})\rvert \rvert_2
    \label{eq:comp}
\end{equation}
Where $\mathrm{NN(\Point{x},\Matrix{P})}$ is the nearest neighbor of a point $\Point{x}$ in a point cloud $\Matrix{P}$. Finally, the overall metric is the mean of accuracy and completeness, also referred to as the Chamfer distance \citep{barrow1977parametric}. All point cloud operations are done in CloudCompare \citep{CloudCompare2025}. 

The reference and estimated point clouds are registered with an Iterative Closest Point (ICP) algorithm \citep{besl1992method} following an automatic coarse registration. In cases of poor initial reconstruction, a manual coarse alignment is occasionally applied to ensure a robust registration. We verify the registration success visually and only report results in case of a registration success. Complete registration failures are usually related to a poor estimated point cloud. Furthermore, the estimated point clouds are cropped automatically, so that no points from the background falsify the metrics. We provide masks for cropping regions of the estimated point cloud, that are not reconstructed in the reference point cloud, as otherwise the accuracy would be falsified. Both VGGT and $\pi^3$ output confidences for the estimated points. For $\pi^3$, we apply the default confidence threshold for cropping points with low confidence. For VGGT, the default confidence cropping threshold of 5 removed too many points, so we empirically chose a threshold of 2.

\begin{figure*}[tbp!]
    \centering
    \scalebox{0.9}{
    \begin{tabular}{@{}cccccc@{\hspace{0.5em}}c@{}}
        \includegraphics[width=0.15\textwidth,trim={0 3cm  0 6cm},clip]{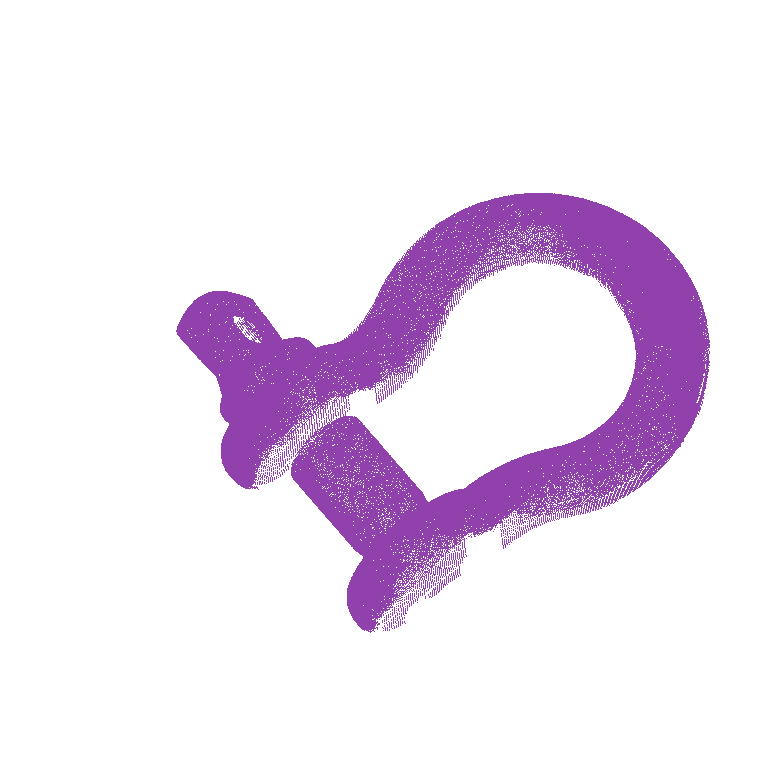} &
        \includegraphics[width=0.15\textwidth,trim={0 3cm  0 6cm},clip]{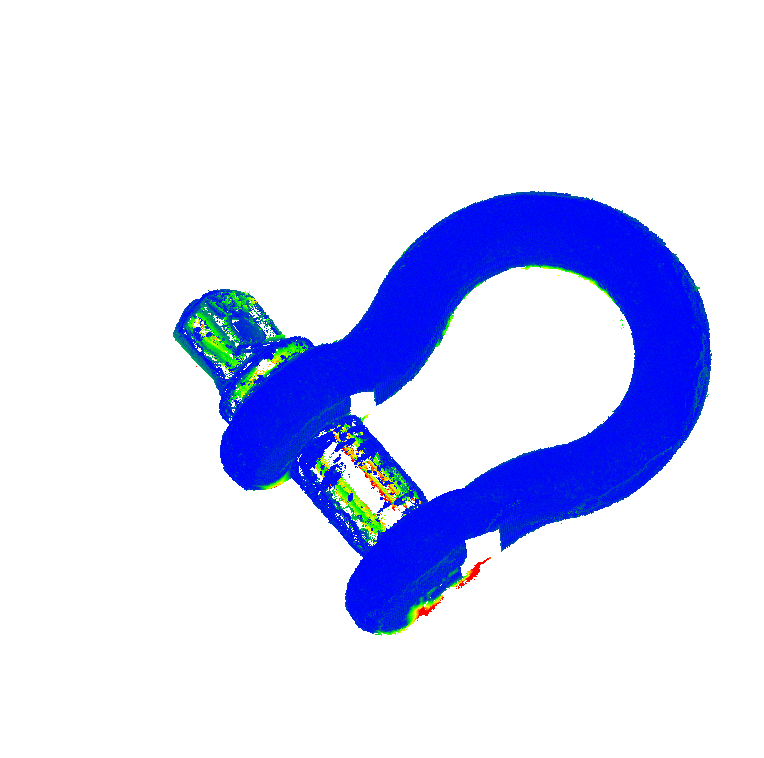} &
        \includegraphics[width=0.15\textwidth,trim={0 3cm  0 6cm},clip]{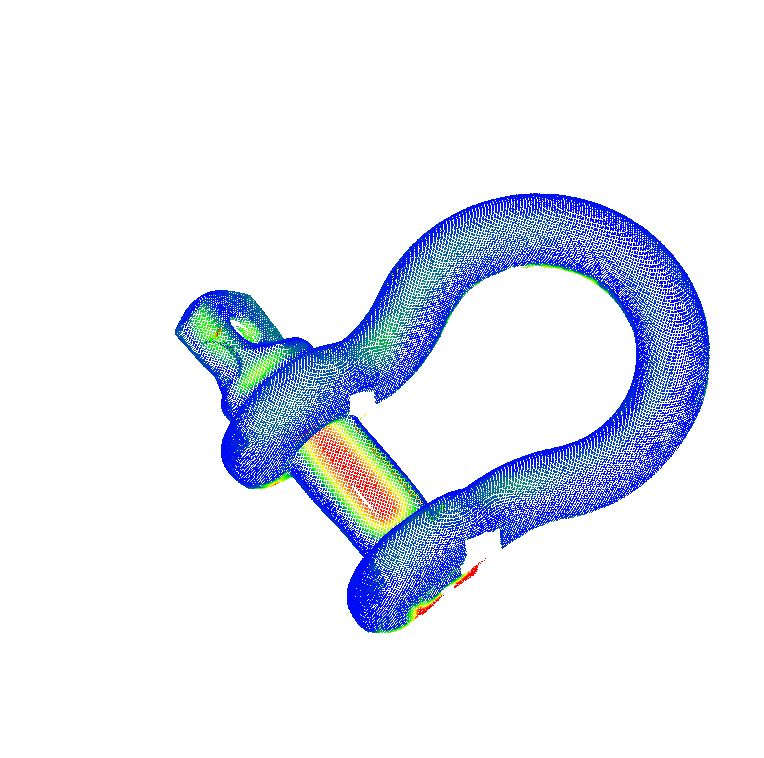} &
        \includegraphics[width=0.15\textwidth,trim={0 3cm  0 6cm},clip]{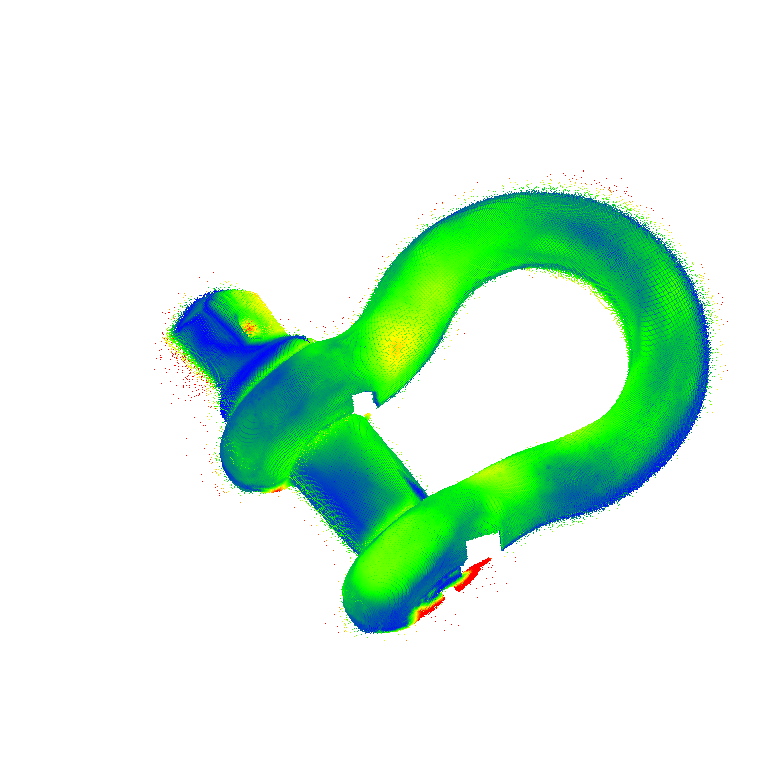} &
        \includegraphics[width=0.15\textwidth,trim={0 3cm  0 6cm},clip]{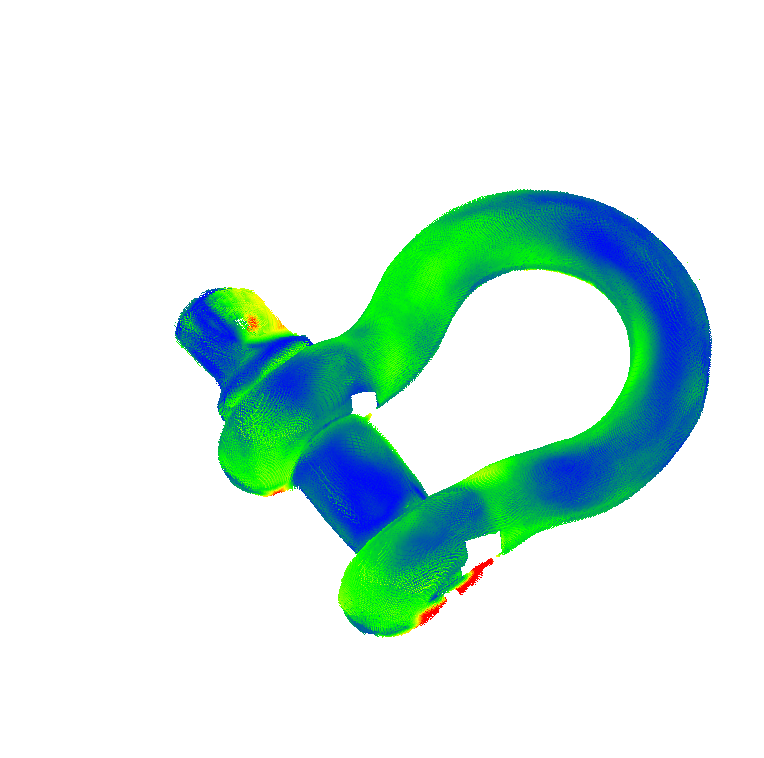} &
        \includegraphics[width=0.15\textwidth,trim={0 3cm  0 6cm},clip]{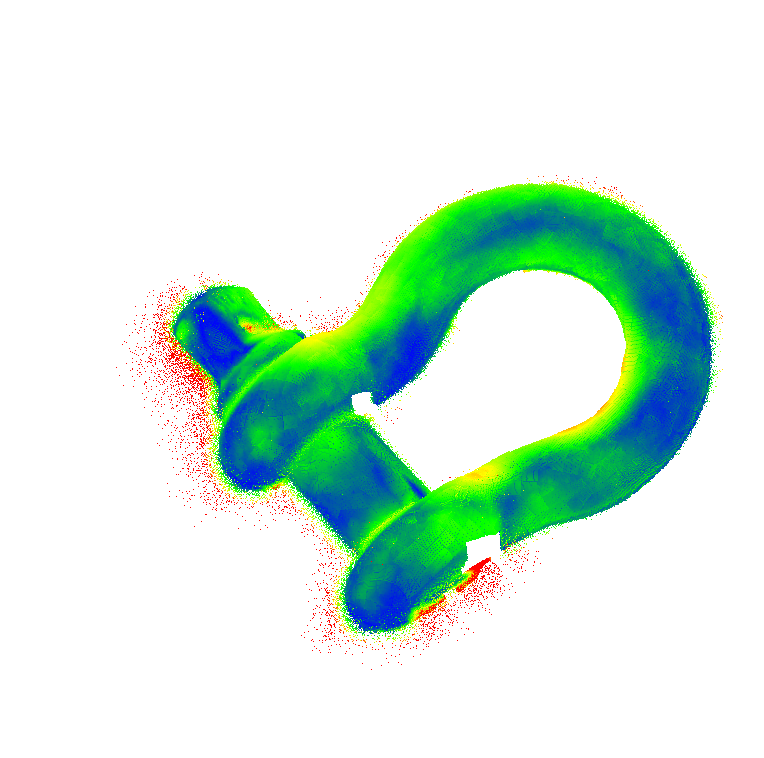} &\\
        \includegraphics[width=0.15\textwidth,trim={0 2cm  0 6cm},clip]{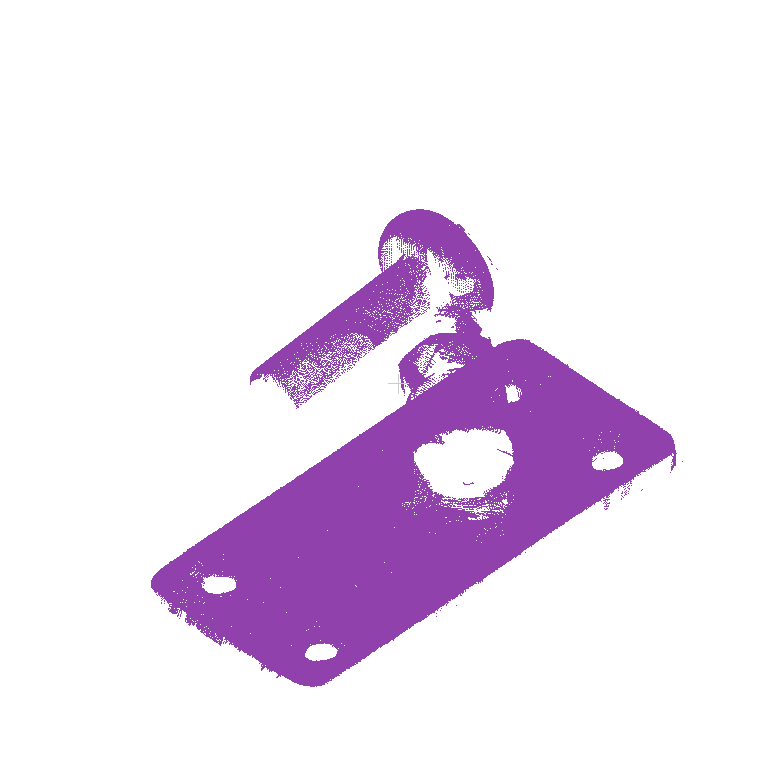} &
        \includegraphics[width=0.15\textwidth,trim={0 2cm  0 6cm},clip]{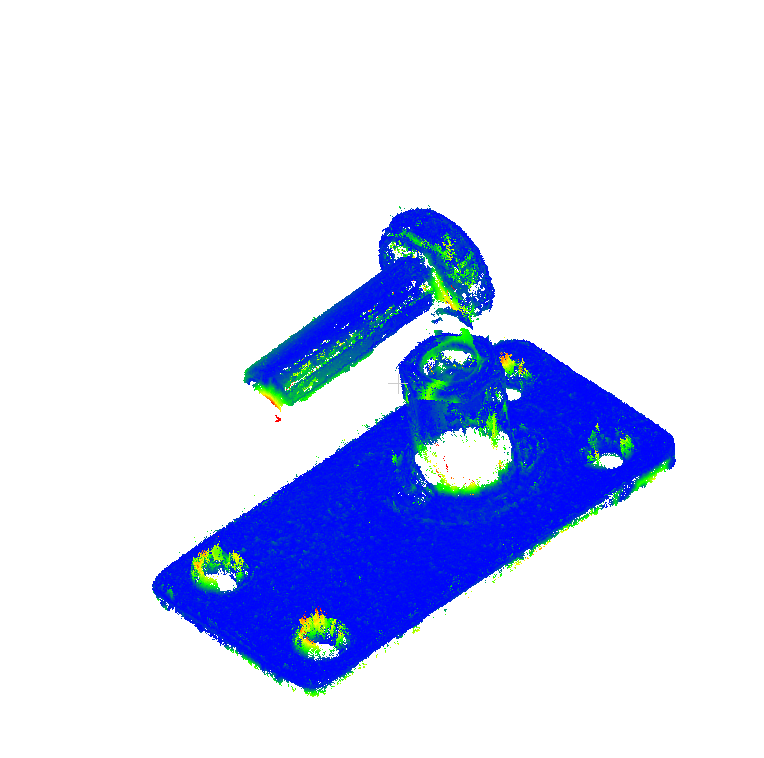} &
        \includegraphics[width=0.15\textwidth,trim={0 2cm  0 6cm},clip]{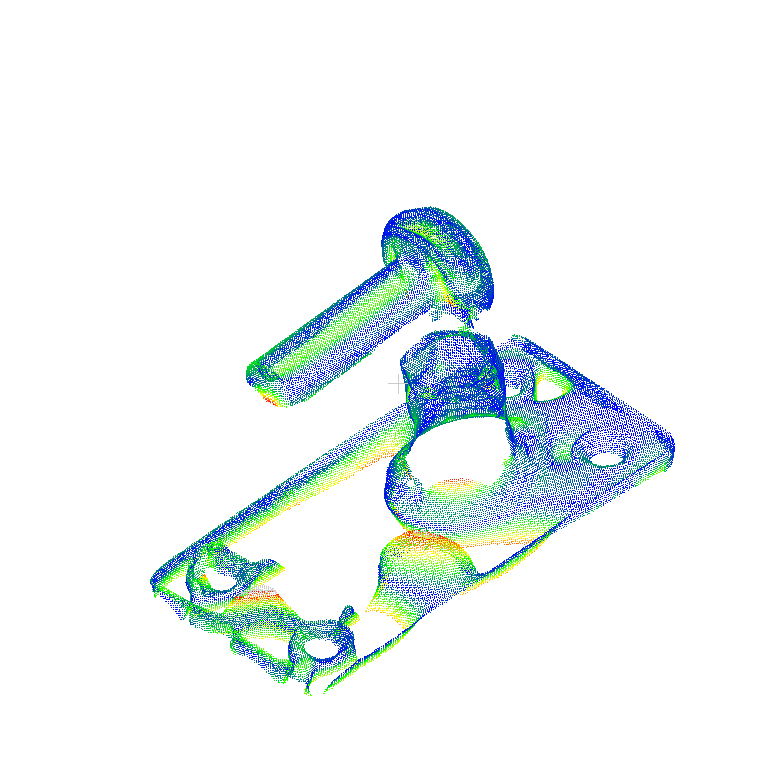} &
        \includegraphics[width=0.15\textwidth,trim={0 2cm  0 6cm},clip]{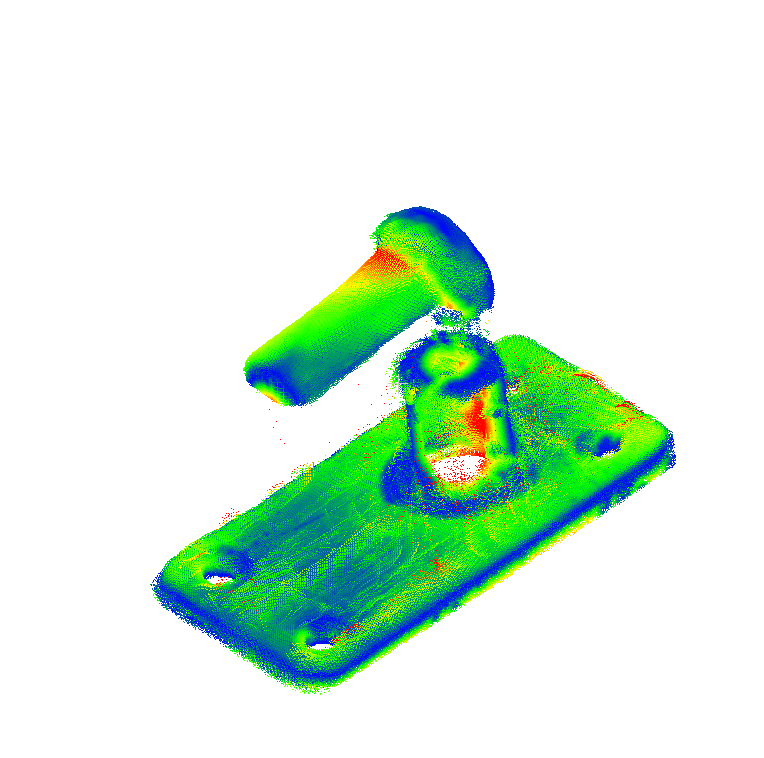} &
        \includegraphics[width=0.15\textwidth,trim={0 2cm  0 6cm},clip]{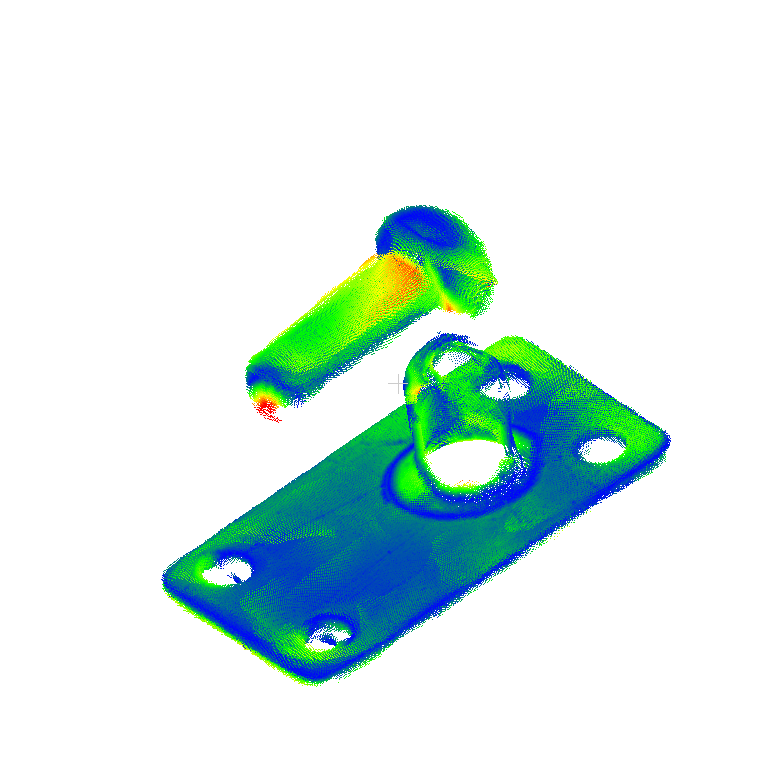} &
        \includegraphics[width=0.15\textwidth,trim={0 2cm  0 6cm},clip]{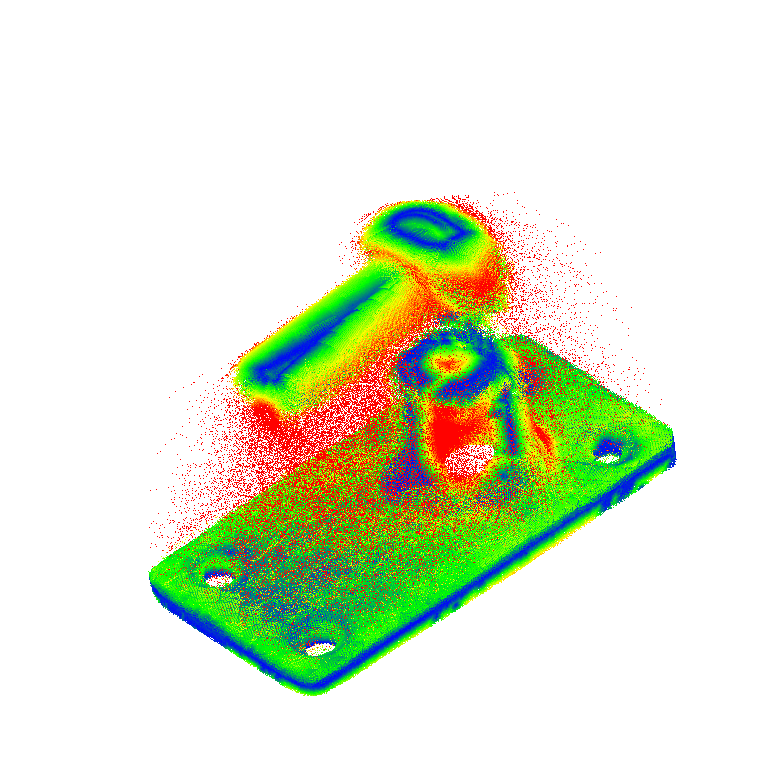} &
\multirow{3}{*}[11em]{
    \raisebox{-0.5\height}{%
        \hspace{-1em}
        \begin{tabular}{c@{\hspace{0.1em}}l}
            \raisebox{0cm}{\includegraphics[height=5.8cm]{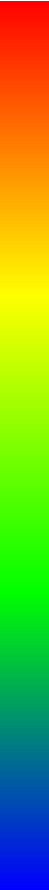}} & 0\,mm \\[-5.8cm]
             & 5\,mm \\
        \end{tabular}%
    }%
}
\\
        \subfloat[Reference]{\includegraphics [width=0.15\textwidth,trim={0 3cm  0 2cm},clip] {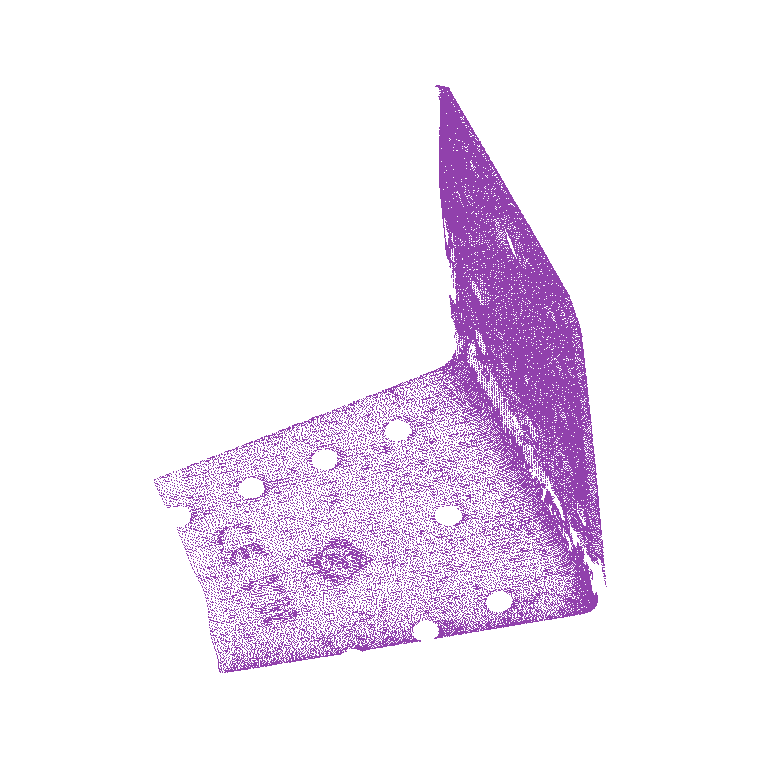}} &
        \subfloat[COLMAP MVS]{\includegraphics[width=0.15\textwidth,trim={0 3cm  0 2cm},clip]{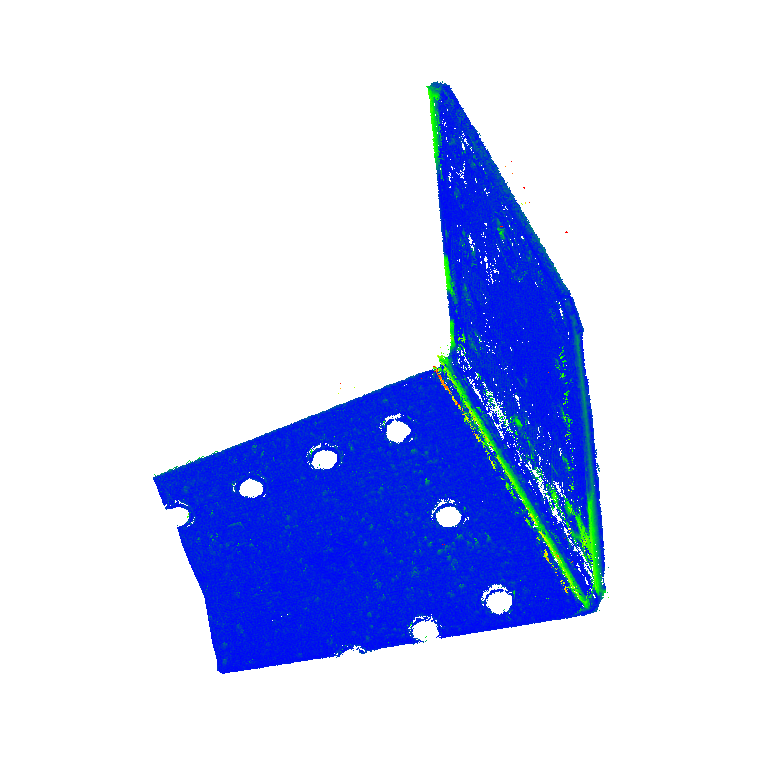}} &
        \subfloat[2DGS]{\includegraphics      [width=0.15\textwidth,trim={0 3cm  0 2cm},clip]{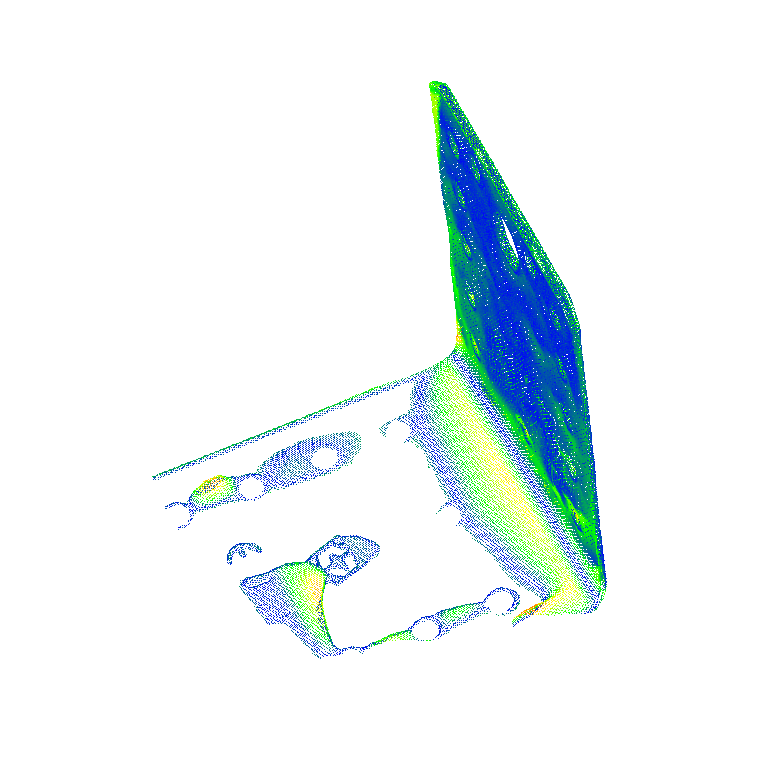}} &
        \subfloat[VGGT depth]{\includegraphics[width=0.15\textwidth,trim={0 3cm  0 2cm},clip]{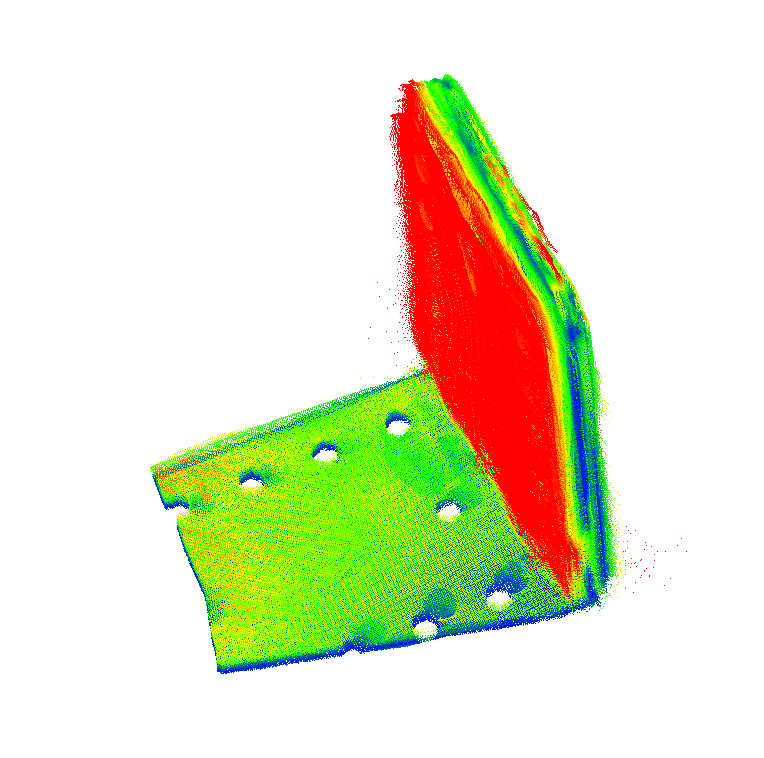}} &
        \subfloat[VGGT point]{\includegraphics[width=0.15\textwidth,trim={0 3cm  0 2cm},clip]{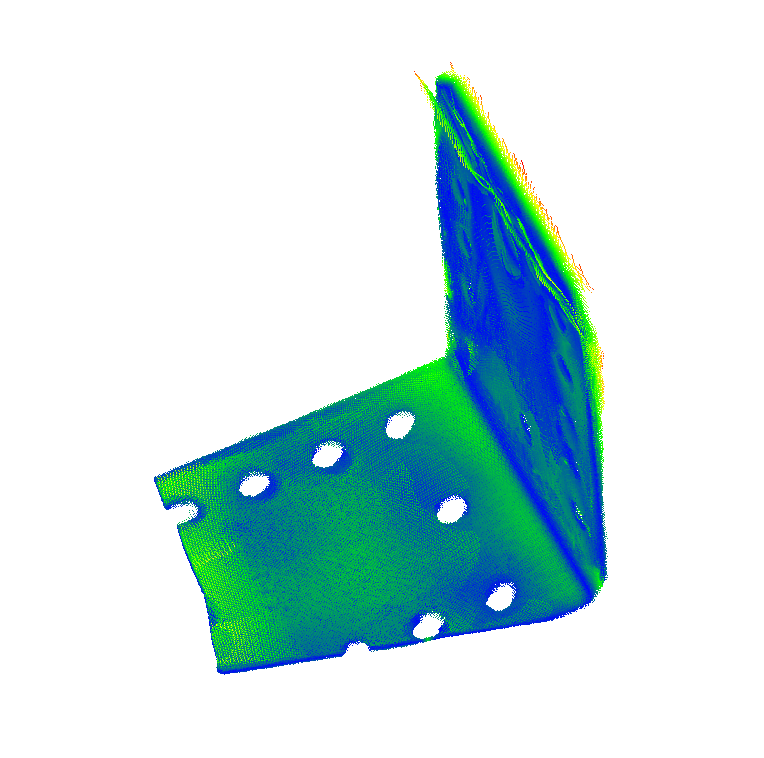}} &
        \subfloat[$\pi^3$]{\includegraphics   [width=0.15\textwidth,trim={0 3cm  0 2cm},clip]{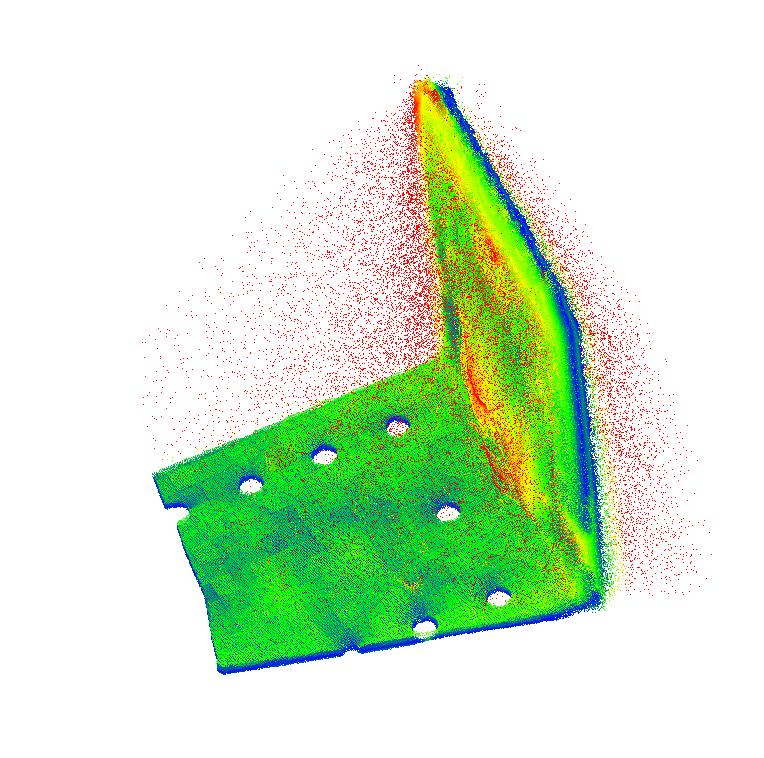}} &
        \\
    \end{tabular}}
    \caption{Pointwise accuracy of the registered point clouds. A saturation threshold of 5\,mm is used for visualization. From top to bottom we show objects 8,6 and 9, all with alu background.}
    \label{fig:accuracy_renders}
\end{figure*}

While the number of points in the VGGT and $\pi^3$ point clouds are determined by the total number of images and their size, the COLMAP MVS and 2DGS reconstruction typically consist of considerably fewer points and are not necessary sampled evenly. Therefore, a spatial subsampling similar to the one used in \cite{aanaes2016large}, but with a different subsampling distance threshold, is applied to $\Matrix{P}_{\mathrm{est}}$ and $\Matrix{P}_{\mathrm{ref}}$.The subsampling distance threshold is set to the scanner accuracy of $0.047$\,mm (see Section \ref{sec:data}). This assures, that regions with high point density do not bias the results. \begin{table*}[tbp!]
\centering
\caption{Mean accuracy, mean completeness and overall in mm for the estimated point clouds of VGGT, $\pi^3$, COLMAP MVS and 2DGS when compared with their respective reference point cloud. For VGGT and $\pi^3$ we use the point clouds generated with the preprocessed images as input. For VGGT, we report the results for the pointcloud generated via the pointmap branch (point) as well as the pointcloud generated via the depthmap branch (depth). Lowest metric for each object and background combination is print bold.}
\label{tab:pointcloud_acc_comp}
\small
\resizebox{\textwidth}{!}{%
    \setlength{\tabcolsep}{10pt}
\begin{tabular}{lccccccccccc}
\midrule
 \textbf{Method} & \textbf{Background}&\textbf{Object 1} & \textbf{Object 2} & \textbf{Object 3} & \textbf{Object 4} & \textbf{Object 5} & \textbf{Object 6} & \textbf{Object 7} & \textbf{Object 8}& \textbf{Object 9}&\textbf{Mean}\\
\midrule
\multicolumn{12}{c}{\textit{Accuracy [mm] $\downarrow$}} \\
\midrule
 VGGT depth&green&3.74&1.25&2.25&1.21&1.29&0.70&1.00&0.70&0.67&1.45\\
 VGGT point&green&\textbf{3.43}&\textbf{0.63}&\textbf{0.70}&\textbf{1.01}&\textbf{0.81}&0.40&\textbf{0.65}&0.43&\textbf{0.29}&0.93\\
 $\pi^3$ &green&-&1.45&2.39&1.79&1.45&1.18&-&0.57&0.66&1.36\\
 COLMAP MVS&green&-&-&-&-&-&\textbf{0.29}&-&\textbf{0.40}&\textbf{0.29}&\textbf{0.33}\\
 2DGS&green&-&-&-&-&-&1.41&-&3.86&4.01&3.09\\
 \midrule
 VGGT depth&alu&7.90&-&-&1.77&1.34&0.84&1.44&0.87&1.27&2.20\\
 VGGT point&alu&\textbf{6.71}&-&-&\textbf{0.57}&0.99&0.53&\textbf{0.60}&0.58&0.50&1.50\\
 $\pi^3$ &alu&-&-&-&-&1.43&1.14&-&0.68&0.86&\textbf{1.02}\\
 COLMAP MVS&alu&9.47&\textbf{0.70}&\textbf{1.88}&0.63&0.80&\textbf{0.27}&0.66&\textbf{0.37}&\textbf{0.25}&1.67\\
 2DGS&alu&6.80&1.16&2.55&1.86&\textbf{0.78}&0.97&1.98&0.55&0.53&1.91\\
\midrule
\multicolumn{12}{c}{\textit{Completeness [mm] $\downarrow$}} \\
\midrule
 VGGT depth&green&\textbf{3.52}&0.48&1.54&0.69&\textbf{0.20}&\textbf{0.11}&\textbf{0.13}&0.12&0.21&0.89\\
 VGGT point&green&5.29&3.63&2.78&1.61&0.30&0.21&0.33&0.15&0.70&1.78\\
 $\pi^3$ &green&-&\textbf{0.45}&\textbf{1.11}&\textbf{0.30}&0.56&\textbf{0.11}&-&\textbf{0.10}&\textbf{0.11}&0.39\\
 COLMAP MVS&green&-&-&-&-&-&\textbf{0.11}&-&0.15&0.52&\textbf{0.26}\\
 2DGS&green&-&-&-&-&-&1.43&-&1.02&1.54&1.33\\
 \midrule
 VGGT depth&alu&7.48&-&-&0.27&\textbf{0.16}&0.12&\textbf{0.12}&0.12&\textbf{0.14}&1.20\\
 VGGT point&alu&7.02&-&-&\textbf{0.18}&0.18&0.32&0.38&0.17&0.27&1.22\\
 $\pi^3$ &alu&-&-&-&-&0.20&0.12&-&\textbf{0.09}&0.18&\textbf{0.15}\\
 COLMAP MVS&alu&\textbf{3.73}&\textbf{0.96}&\textbf{2.66}&0.43&0.89&\textbf{0.11}&0.25&0.14&0.16&1.04\\
 2DGS&alu&6.03&1.31&3.02&1.57&0.66&1.73&0.85&0.50&1.00&1.85\\
\midrule
\multicolumn{12}{c}{\textit{Overall (Chamfer distance) [mm] $\downarrow$}} \\
\midrule
 VGGT depth&green&\textbf{3.63}&\textbf{0.87}&1.89&\textbf{0.95}&0.75&0.41&0.56&0.41&0.44&1.17\\
 VGGT point&green&4.36&2.13&\textbf{1.74}&1.31&\textbf{0.55}&0.31&\textbf{0.49}&0.29&0.49&1.35\\
 $\pi^3$ &green&-&0.95&1.75&1.04&1.01&0.65&-&0.34&\textbf{0.39}&0.87\\
 COLMAP MVS&green&-&-&-&-&-&\textbf{0.20}&-&\textbf{0.28}&0.41&\textbf{0.30}\\
 2DGS&green&-&-&-&-&-&1.42&-&2.44&2.77&2.21\\
 \midrule
 VGGT depth&alu&7.69&-&-&1.02&0.75&0.48&0.78&0.50&0.71&1.70\\
 VGGT point&alu&6.86&-&-&\textbf{0.37}&\textbf{0.59}&0.42&0.49&0.38&0.39&1.36\\
 $\pi^3$ &alu&-&-&-&-&0.82&0.63&-&0.39&0.52&\textbf{0.59}\\
 COLMAP MVS&alu&6.60&\textbf{0.83}&\textbf{2.27}&0.53&0.84&\textbf{0.19}&\textbf{0.46}&\textbf{0.25}&\textbf{0.20}&1.35\\
 2DGS&alu&\textbf{6.42}&1.24&2.79&1.72&0.72&1.35&1.42&0.53&0.76&1.88\\
\midrule
\end{tabular}
}\end{table*} Table \ref{tab:pointcloud_acc_comp} shows the mean accuracy, mean completeness, and overall metric for all objects. As in \cite{aanaes2016large}, we disregard observations beyond a threshold of 20\,mm, which we consider outliers. As we have shown in the camera pose evaluation of Section \ref{sec:Pose Comparison}, the preprocessing steps of Section \ref{sec:Image preprocessing} lead to better camera poses. Figure \ref{fig:point_cloud_fail} exemplarily shows that this applies to the point clouds as well. Therefore, for VGGT and $\pi^3$ we only report the results for the preprocessed images, as in most cases even the registration is failing with the original images. For both VGGT and $\pi^3$, the object and background combinations that resulted in bad camera poses also tend to lead to either failed registrations or high accuracy and completeness metrics (compare Tables \ref{tab:pose_acc_combined} and \ref{tab:pointcloud_acc_comp}), which makes sense as they are estimated simultaneously. VGGT point and COLMAP MVS achieve mostly submillimeter accuracy with both background choices, given that the registration worked. $\pi^3$, 2DGS, and VGGT depth perform slightly worse. The worst completeness results are reported for 2DGS, which is mainly caused by erroneous depth estimation on textureless metallic surfaces. This issue can be seen in Figure \ref{fig:accuracy_renders}.\begin{table*}[h!]
\centering
\caption{Mean Rotation Error (MRE) and Mean Translation Error (MTE) for VGGT and $\pi^3$ on the NeRF Synthetic dataset. All experiments are done with a subsampling of $n=100$ images. The results for the original images, objects naturally facing up, are shown in black. In parentheses, we report the results for the rotated images, objects facing down and mark them in green if the rotation improved the result and in red if the opposite is the case. Lowest metric for each object is print bold.}
\label{tab:ablation_study}
\small 
\begin{adjustbox}{width=\textwidth,center}
\begin{tabular}{cccccccccc}
\midrule
 \textbf{Method} & \textbf{chair} & \textbf{drums} & \textbf{ficus} & \textbf{hotdog} & \textbf{lego} & \textbf{materials} & \textbf{mic} & \textbf{ship} &\textbf{Mean}\\
\midrule
\multicolumn{10}{c}{\textit{Mean Rotation Error (MRE) [\textdegree] $\downarrow$}} \\
\midrule
 VGGT & 0.67 (\textcolor{green!60!black}{0.63}) & \textbf{0.23} (\textcolor{red}{1.01}) & 3.01 (\textcolor{green!60!black}{2.94}) & \textbf{0.41} (\textcolor{red}{0.68}) & \textbf{0.40} (\textcolor{red}{1.05}) & 1.67 (\textcolor{red}{1.87}) & 0.86 (\textcolor{red}{1.30}) & \textbf{0.42} (\textcolor{red}{2.45})& 0.96 (\textcolor{red}{1.49}) \\
 $\pi^3$ & \textbf{0.57} (\textcolor{red}{1.22}) & 0.44 (\textcolor{red}{1.37}) & \textbf{2.41} (\textcolor{red}{3.99}) & 0.60 (\textcolor{red}{1.76}) & 0.47 (\textcolor{red}{1.23}) & \textbf{0.72} (\textcolor{red}{1.02}) & \textbf{0.59} (\textcolor{red}{1.16}) & 0.50 (\textcolor{red}{6.83}) &\textbf{0.79} (\textcolor{red}{2.32})\\
\midrule
\multicolumn{10}{c}{\textit{Mean Translation Error (MTE) [mm] $\downarrow$}} \\
\midrule
 VGGT & \textbf{33.17} (\textcolor{red}{44.40}) & \textbf{13.42} (\textcolor{red}{59.97}) & 199.83 (\textcolor{green!60!black}{188.28}) & \textbf{21.34} (\textcolor{red}{41.64}) & \textbf{16.73} (\textcolor{red}{62.29}) & 103.04 (\textcolor{green!60!black}{101.75}) & 44.35 (\textcolor{red}{64.24}) & \textbf{25.11} (\textcolor{red}{210.16})& \textbf{56.62} (\textcolor{red}{96.59})\\
 $\pi^3$ & 48.78 (\textcolor{red}{78.61}) & 26.55 (\textcolor{red}{82.51}) & \textbf{122.83} (\textcolor{red}{164.30}) & 76.39 (\textcolor{red}{97.93}) & 25.10 (\textcolor{red}{87.25}) & \textbf{71.27} (\textcolor{red}{83.51}) & \textbf{41.95} (\textcolor{red}{81.91}) & 42.90 (\textcolor{red}{152.63})& 56.97 (\textcolor{red}{103.58})\\
\midrule
\end{tabular}
\end{adjustbox}
\end{table*} The bad completeness in combination with the long training of 2DGS appears to be not worth the effort, except if the generated novel views are of interest. $\pi^3$ along with COLMAP MVS generally tend to achieve the best completeness values. The main reason for VGGT point falling behind in this regard, is that the confidences of the point branch appear to be lower than those of the depth branch, resulting in more points cropped. Regarding the overall metric, VGGT depth/point, $\pi^3$, and COLMAP MVS are not significantly different in the big picture, though exceptions exist (e.g., for object 6 alu, COLMAP MVS is significantly better). Taking the processing times of VGGT and $\pi^3$ into account, the results are remarkable.

\subsection{Ablation Study: NeRF Synthethic}
\label{sec:Ablation Study}
We verify our findings that VGGT as well as $\pi^3$ are sensitive to the rotation of input images. We use a subset of $n=100$ images of the NeRF Synthetic dataset \citep{mildenhall2021nerf} and report the MRE  (Equation~(\ref{eq:MRE})) and MTE (Equation~(\ref{eq:MTE})) for the original images and for the images rotated by $180^\circ$ in Table \ref{tab:ablation_study}. The MTE and MRE values cannot be compared directly to the MTE and MRE values of MVM-IOD or to the other scenes within the NeRF synthetic dataset, as each scene has a different extent. However, if VGGT and $\pi^3$ were invariant regarding the rotation of the input images, one would expect similar results for the MRE and MTE, independent of the rotation. While this is the case for the scenes materials and ficus, where the MRE and MTE values do not change significantly, the MRE and MTE do worsen significantly for the other scenes. In most cases, the $\pi^3$ camera poses are significantly more affected by the rotation than the VGGT camera poses.

\section{Conclusion}\label{sec:Conclusion}We present MVM-IOD, a dataset for the evaluation of camera pose estimation and 3D reconstruction on industrial objects. We show, that MVM-IOD is highly relevant, as novel feed-forward geometry transformers struggle with these type of input images more than with non-industrial data. While the tested 3D reconstruction methods do not generate accurate 3D reconstructions for some objects in the MVM-IOD, we see this as a challenge for future methods, as our motivation is to provide an extremely challenging dataset. We also empirically show that VGGT and $\pi^3$ have some sort of scene gravity awareness, as the results worsen when the scene contents seem upside down. This means, that in industrial scenarios, the operator should be aware that current feed-forward geometry transformers can fail, when the input images are lying too far out of distribution. This is especially relevant, as the feed-forward methods will always produce an output unlike other tested methods like COLMAP, which simply fails in adverse conditions. The images can be shifted closer to the distribution by applying preprocessing steps to the images. Or, these findings can be used to train or finetune new feed-forward geometry transformers with upside down images and images of varying aspect ratios.

{	\begin{spacing}{1.17}
		\normalsize
		\bibliography{main} 
	\end{spacing}}
\end{document}